%% file: IEEE_journal.tex
\documentclass[journal,twoside]{IEEEtran}

\input{common/setup.tex}
\usepackage{orcidlink}
\hypersetup{
  hidelinks,
  pdftitle={Mixture of Flow Matching and Particle Filter for Proprioception-Based Online Contact Estimation},
  pdfauthor={Seo Wook Han and Min Jun Kim},
  pdfsubject={Author Accepted Manuscript; IEEE Transactions on Robotics, vol. 42, pp. 3646-3665, 2026; DOI: 10.1109/TRO.2026.3723975}
}

\title{Mixture of Flow Matching and Particle Filter for Proprioception-Based Online Contact Estimation}

\author{Seo Wook Han$^{\orcidlink{0009-0007-8351-2368}}$,
and Min Jun Kim$^{\orcidlink{0000-0003-0605-4512}}$, \IEEEmembership{Member,~IEEE,}
\thanks{
Manuscript received: December 19, 2025; Revised: June 4, 2026; Accepted: July 13, 2026.
This work was supported by 
the National Research Foundation of Korea (NRF) Grant funded by the Korean government (MSIT) (Project RS2025-02213804), 
the Industrial Technology Innovation Program funded by the Ministry of Trade, Industry and Resources (MOTIR) through the Korea Evaluation Institute of Industrial Technology (KEIT) (Grant No. RS-2024-00441872) and the Korea Institute for Advancement of Technology (KIAT) (Grant No. P0028404), 
and BK21 FOUR (Connected AI Education \& Research Program for Industry and Society Innovation, KAIST EE, No. 4120200113769). 
(Corresponding author: Min Jun Kim.)

Seo Wook Han and Min Jun Kim are with the Intelligent Robotic Systems Laboratory,
School of Electrical Engineering, Korea Advanced Institute of Science and Technology
(KAIST), Daejeon 34141, South Korea
(e-mail: \texttt{seowook.han@kaist.ac.kr; minjun.kim@kaist.ac.kr}).

}}
\makeatletter
\def\ps@aam{%
  \def\@oddhead{\hbox{}\@IEEEheaderstyle \MakeUppercase{Han and Kim: Mixture of Flow Matching and Particle Filter for Proprioception-Based Online Contact Estimation}\hfil}%
  \def\@evenhead{\hfil\@IEEEheaderstyle IEEE TRANSACTIONS ON ROBOTICS, 2026\hbox{}}%
  \let\@oddfoot\@empty
  \let\@evenfoot\@empty
}
\let\ps@IEEEtitlepagestyle\ps@aam
\makeatother
\begin{document}

\begin{titlepage}
\pagenumbering{roman}
\centering
\vspace*{0.5in}
{\Large\bfseries Author Accepted Manuscript (AAM)\par}
\vspace{2\baselineskip}
{\LARGE Mixture of Flow Matching and Particle Filter for Proprioception-Based Online Contact Estimation\par}
\vspace{1.5\baselineskip}
{\large Seo Wook Han and Min Jun Kim\par}
\vspace{3\baselineskip}
\begin{minipage}{0.9\textwidth}
\normalsize\raggedright
This is the final author manuscript incorporating the revisions made during peer review.\par
\medskip
Published in \emph{IEEE Transactions on Robotics}, vol.~42, pp.~3646--3665, 2026.\par
\medskip
The version of record is available at\par
\href{https://doi.org/10.1109/TRO.2026.3723975}{https://doi.org/10.1109/TRO.2026.3723975}.\par
\bigskip
\copyright~2026 IEEE. Personal use of this material is permitted. Permission from IEEE must be obtained for all other uses, in any current or future media, including reprinting/republishing this material for advertising or promotional purposes, creating new collective works, for resale or redistribution to servers or lists, or reuse of any copyrighted component of this work in other works.\par
\end{minipage}
\vfill
\end{titlepage}
\pagenumbering{arabic}

\maketitle

\input{abstract.tex}

\input{core_text.tex}

\bibliographystyle{common/myIEEEtran}
\bibliography{common/IEEEabrv,common/references}

\end{document}

%% file: common/setup.tex
\usepackage{graphicx}
\usepackage{amsmath}
\usepackage{arydshln}
\usepackage{hhline}
\usepackage{eucal}
\usepackage{cite} 
\usepackage{amssymb}
\usepackage{enumerate}
\usepackage[caption=false,font=footnotesize]{subfig}
\usepackage{bm}
\usepackage{color}
\usepackage{multirow}
\usepackage{algorithm} 
\usepackage{algpseudocode}

\usepackage{booktabs}

\usepackage{mathtools}

\newtheorem{rem}{Remark}

\newtheorem{assumption}{Assumption}
\providecommand{\norm}[1]{\lVert#1\rVert}

\makeatletter
\renewcommand{\env@matrix}[1][c]{%
  \hskip-\arraycolsep
  \let\@ifnextchar\new@ifnextchar
  \array{*\c@MaxMatrixCols #1}%
}
\makeatother

\DeclareSymbolFontAlphabet{\mathcal}   {symbols}

\usepackage{tikz}
\usetikzlibrary{calc}
\newcommand*\circled[1]{\tikz[baseline=(char.base)]{
    \node[shape=circle, draw, inner sep=1pt, 
        minimum height={\f@size*1.6},] (char) {\vphantom{WAH1g}#1};}}
\makeatother

\algrenewcommand{\algorithmiccomment}[1]{\hfill{\scriptsize\(\triangleright\) \texttt{#1}}}
\newcommand{\codeComment}[1]{\Comment{\texttt{\scriptsize #1}}}
\newcommand{\preComment}[1]{\Statex {\scriptsize\texttt{// #1}}}
\usepackage{tabularx} 
\usepackage{pifont}

\usepackage{refcount} 
\makeatletter
\newcommand{\subsecref}[1]{%
  \expandafter\@secondoftwo\csname r@#1\endcsname}
\makeatother

\usepackage{physics} 

\usepackage{siunitx}    
\usepackage{float}

\usepackage{enumitem}  

%% file: abstract.tex
\begin{abstract}
In this article, we propose ContactFlowFilter, a novel approach for contact estimation that determines the number of active contacts, their locations, and the associated force vectors. 
We consider a robot manipulator equipped with joint torque sensors and a base force/torque sensor; such a proprioceptive-only setup presents a challenging inverse problem characterized by (i) multi-modality from non-unique solutions, (ii) temporal dependencies in state evolution, and (iii) discrete contact events involving contact addition/removal.
Existing approaches, such as particle filters (PFs) and learning-based methods, often struggle to address these challenges simultaneously.
To overcome this, ContactFlowFilter integrates a generative model into the PF framework.
The recursive structure of the PF captures temporal dependencies, while the flow matching model serves as an auxiliary proposal distribution to track multi-modal beliefs and detect contact events.
Extensive validation in both simulation and real-world experiments confirms the effectiveness of the proposed method. 
In real-world settings, ContactFlowFilter achieves the lowest dissimilarity from the target distribution, with errors of only 0.36~cm and 0.88~cm for single- and dual-contact scenarios, respectively, while operating at 0.69~ms and 2.69~ms.
\end{abstract}


%% file: core_text.tex
\newcommand{\pidx}[1]{^{(#1)}}
\newcommand{\fidx}[1]{^{[#1]}}
\newcommand{\fra}[1]{{\{#1\}}}
\newcommand{\expgenerative}{A}
\newcommand{\expmanagepset}{B}
\newcommand{\expbelief}{C}
\newcommand{\expreal}{D}

\begin{figure}[t!]
\centering
	\subfloat[]
    {\includegraphics[width=0.5\columnwidth, trim = {2px 0px 0 0px}, clip]{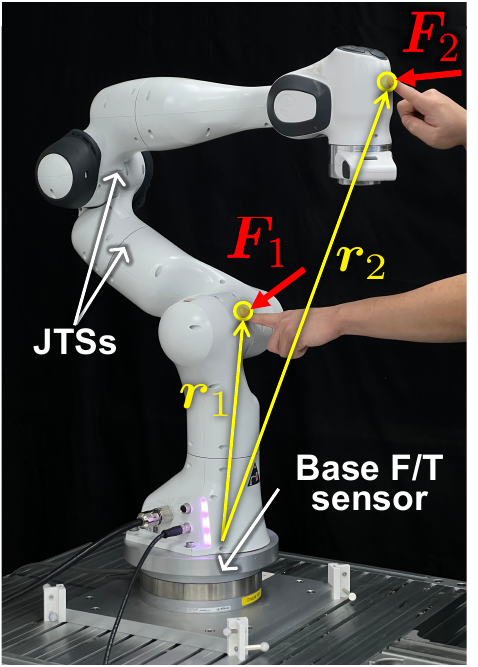}}
    \subfloat[]
    {\includegraphics[width=0.5\columnwidth, trim = {2px 0px 0 0px}, clip]{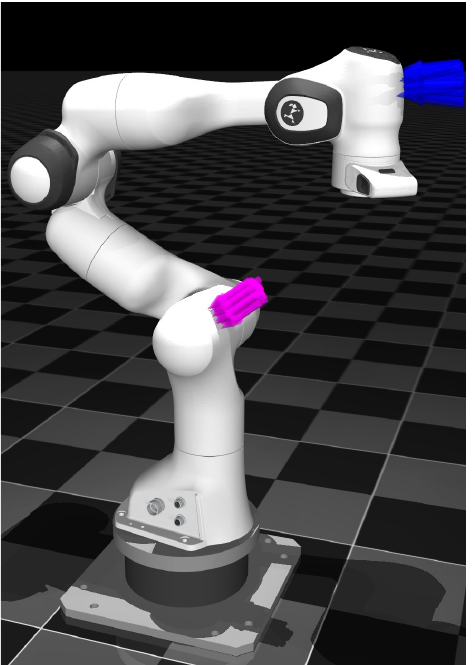}}
	\caption{
        (a) This work focuses on the problem of contact estimation for robotic arms: identifying the number of active contacts, their locations ($\bm{r}_{i} \in \mathbb{R}^3$), and the associated forces ($\bm{F}_{i} \in \mathbb{R}^3$) using proprioceptive sensing (e.g., joint torque sensors (JTSs) and a base force/torque (F/T) sensor). 
        (b) The proposed method addresses up to two active contacts, with the estimation results for the first and second contacts represented by magenta and blue arrows, respectively.
	}
	\label{fig:intro_figure}
    \vspace{-3mm}
\end{figure}

\section{Introduction}
\label{sec:introduction}
\IEEEPARstart{C}{ontact} information is crucial for robots performing \textit{physical interactions}.
While collision detection \cite{8932392, heo2019collision, kim2021transferable} typically provides simple binary information (i.e., contact or no-contact), estimating a detailed contact state that includes contact location and force (e.g., $\bm{r}_i$ and $\bm{F}_i$ in Fig.~\ref{fig:intro_figure}) unlocks a wide range of interactive applications.
For example, such detailed contact information can transform the robot's body into an intuitive human-robot interface, allowing users to trigger different commands by touching specific locations \cite{fu2025unitac, iskandar2024intrinsic} or to control the robot by leveraging the contact state as a virtual joystick \cite{confente2025skin}.
It also supports interactive control schemes, such as constraining the contact force on the robot's body \cite{pang2022easing}, and substituting for a dedicated force/torque (F/T) sensor used for admittance \cite{kim2016passivity, kim2019passivity, 10637775} or direct force control \cite{10107744}. 
Furthermore, this detailed estimation capability enables advanced strategies such as hybrid motion-force control under unknown forces from outside the target region \cite{10611151}, safe navigation through cluttered environments where unexpected contacts can occur \cite{jain2013reaching, killpack2016model, saleem2025contact}, and whole-body manipulation of humanoid robots \cite{11037531, barreiros2025learning, armleder2025real}.

\begin{figure*}[t!]
\centering
    \subfloat[\scriptsize Multi-Modality]
	{\includegraphics[width=0.262\textwidth, trim = {0 0px 0 0px}, clip]{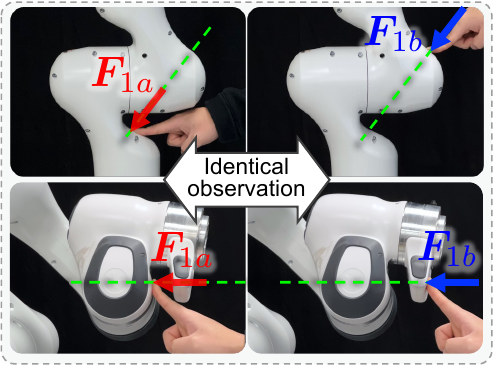}}
    \subfloat[{\scriptsize Temporal Dependency}]
	{\includegraphics[width=0.424\textwidth, trim = {0 0px 0 0px}, clip]{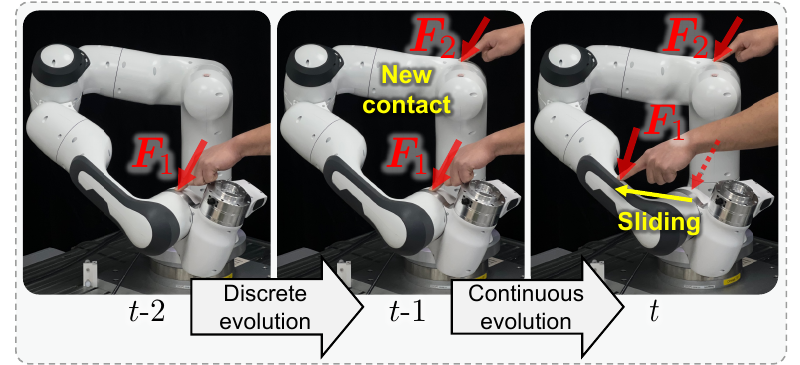}}
    \subfloat[{\scriptsize Contact Event Detection}]
	{\includegraphics[width=0.314\textwidth, trim = {0 0px 0 0px}, clip]{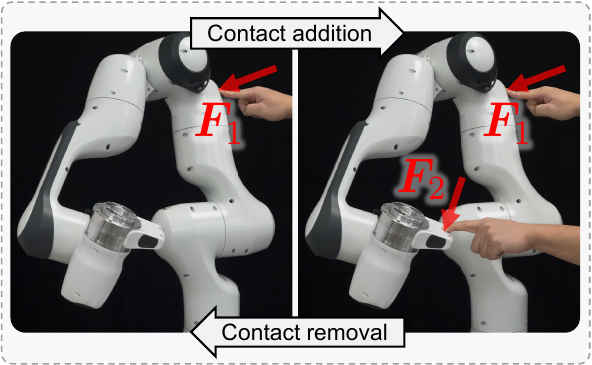}}
	\caption{
    Examples of key challenges in the contact estimation problem:
    (a) \textit{Multi-modality}: Two distinct contact forces ($\bm{F}_{1a}$ and $\bm{F}_{1b}$) result in identical observations.
    (b) \textit{Temporal dependency}: Current contact states depend on history, involving discrete changes in the number of contacts (e.g., when a new contact $\bm{F}_2$ occurs, $t\text{-}2\!\rightarrow \!t\text{-}1$) and continuous updates in locations or forces (e.g., when an existing contact $\bm{F}_1$ slides, $t\text{-}1\!\rightarrow \!t$).
    (c) \textit{Contact event detection}: 
    Determining the number of active contacts relies on identifying the exact timing of a transition and whether a contact is added or removed.
	}
	\label{fig:challenges}
    \vspace{-3mm}
\end{figure*}

In essence, all these applications rely on the ability to identify the detailed contact state, which consists of (i) the total number of contacts, (ii) their locations, and (iii) the associated force vectors.
However, accurately estimating the contact state remains a largely unsolved problem.
To address this, researchers have explored two main sensory pathways. 
The first approach uses exteroceptive sensors, such as surface-mounted tactile skins \cite{mittendorfer2015realizing, ho2021iotouch}, to directly obtain rich contact information at the interaction region. 
The second relies on proprioceptive sensors, such as joint torque sensors (JTSs) and F/T sensors \cite{cdm, manuelli2016localizing, iskandar2021collision}, to infer the contact state from the robot's internal state. 
We focus on the latter approach because these sensors are widely available and integrated into modern robotic systems. 
For instance, commercially available platforms such as the Franka Emika Panda, KUKA iiwa and iisy, and Kinova Gen3 are already equipped with JTSs, while F/T sensors can be readily installed (see Fig.~\ref{fig:intro_figure}).

However, due to their sparse placement, proprioceptive sensors provide only indirect information about external contacts, as they capture the resulting internal joint torques and forces.
We need to solve an inverse problem that estimates the contact state that best explain these sparse, internal sensor measurements.
As shown in Fig.~\ref{fig:intro_figure}, our objective is to estimate the contact state, which includes the contact locations on distinct links (${\bm{r}}_i\!\in\!\mathbb{R}^3$), the corresponding three-dimensional (3D) force vectors (${\bm{F}}_i\!\in\!\mathbb{R}^3$), and the number of contacts (up to two). 
By constraining our scope to scenarios with up to two point contacts, we isolate and tackle the core challenges of the contact estimation problem:
\begin{itemize}[leftmargin=*, nosep]
    \item \textit{\textbf{Multi-modality}} (\textit{Singularity}):
    The primary challenge stems from the inherent ambiguity of the inverse problem. 
    While the forward mapping (from contact to sensor reading) is deterministic, the inverse mapping is ambiguous; as illustrated in Fig.~\ref{fig:challenges}(a), distinct contact states may correspond to identical sensor measurements.
    This \textit{singularity}~\cite{pang2021identifying, popov2021real} implies that the contact state might not be reliably determined from a single observation, leading to inherent uncertainty in the estimation.
    Consequently, the probability distribution of possible solutions becomes \textit{multi-modal}.
    Due to the uncertain nature of the problem, it is more informative and important to predict the \textit{multi-modal} distribution, rather than yielding a single deterministic solution.
    
    \item \textit{\textbf{Temporal Dependency}}:
    Although the ambiguity of inferring contact states from single observations remains an inherent problem, it can be mitigated in specific contexts by leveraging \textit{temporal dependency}. 
    For instance, consider the discrete state evolution depicted in Fig.~\ref{fig:challenges}(b). 
    In this scenario, the ambiguity is significantly compounded as the number of contacts increases (e.g., when simultaneously estimating two contacts). 
    Nevertheless, estimating the second contact becomes more reliable by incorporating past information about the first contact \cite{cdm}.
    In the case of continuous evolution such as the sliding contact force shown in Fig.~\ref{fig:challenges}(b), knowing the previous contact location simplifies the estimation.
    Performing localization around the previous estimate improves efficiency by avoiding the need to explore the entire robot surface \cite{manuelli2016localizing}.
    Consequently, leveraging \textit{temporal dependency}, specifically utilizing past estimates, provides a prior context for inferring the current state.
    
    \item \textit{\textbf{Contact Event Detection}}:
    Determining the number of active contacts requires identifying discrete transitions, specifically the \textit{addition} or \textit{removal} of contacts (see Fig.~\ref{fig:challenges}(c)).
    While providing a prior context enables effective temporal tracking, its validity strictly depends on the correct discrete state; otherwise, undetected contact events cause the estimator to propagate incorrect past information.
    Therefore, robust \textit{contact event detection} serves as a critical prerequisite for reliable estimation of the complete contact state, including the number, location, and force.
\end{itemize}
Furthermore, these challenges need to be addressed while remaining computationally efficient, as practical applications often require real-time estimation of the contact state.

As detailed in Section~\ref{sec:related_works}, prior work on proprioception-based contact estimation can be broadly categorized into model-based and learning-based approaches. 
The former is dominated by particle filter (PF)-based methods \cite{manuelli2016localizing, 10161173, zwiener2019armcl, popov2021multi, job2023multiple}, while the latter includes deep learning techniques, from probabilistic mapping \cite{liang2021contact} to diffusion models \cite{cdm}. Each of these categories exhibits distinct strengths and weaknesses.

\begin{figure*}[t!]
\centering
	{\includegraphics[width=0.95\textwidth, trim = {0 0px 0 0px}, clip]{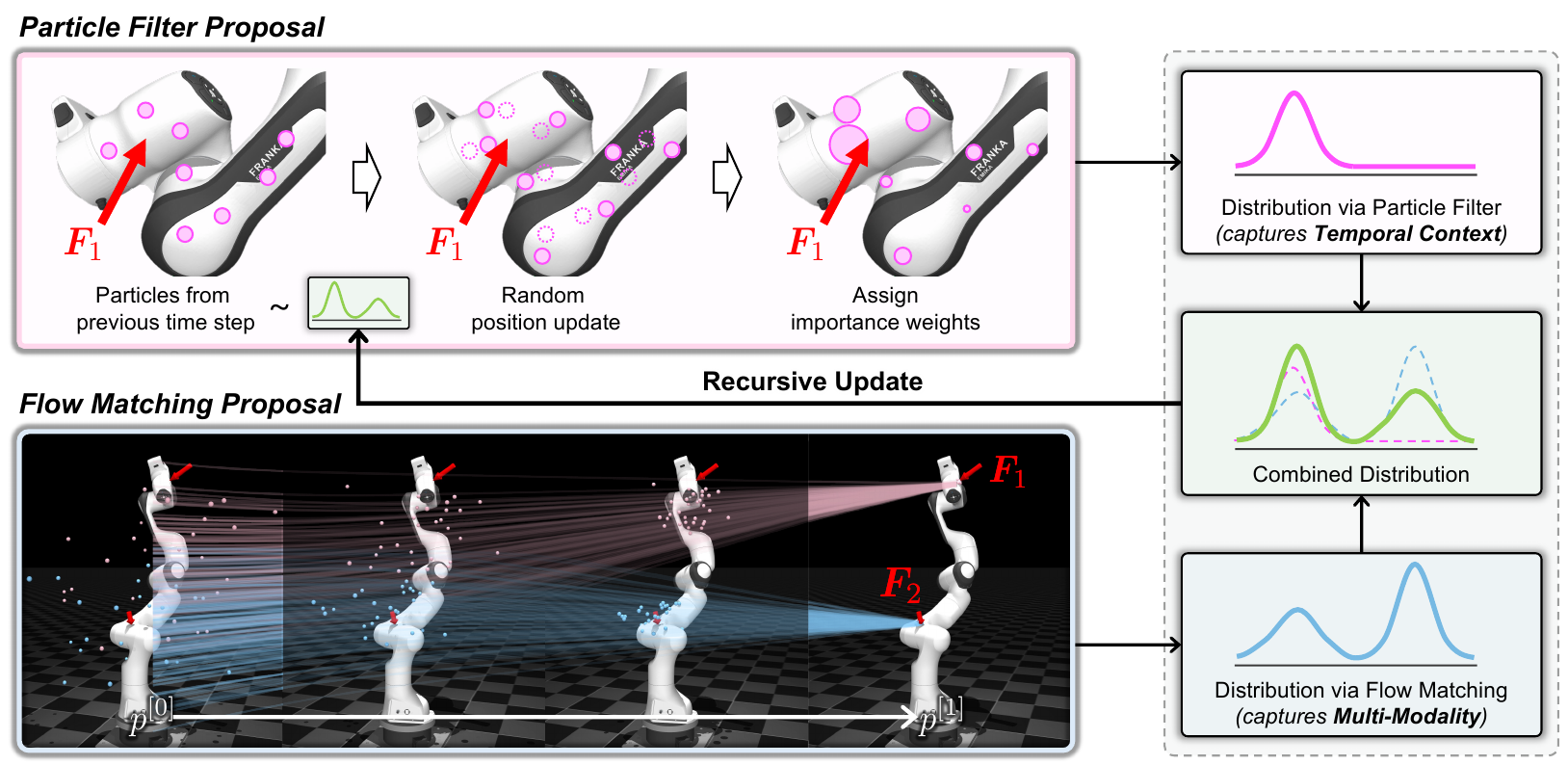}}
    \vspace{-3mm}
	\caption{
        Overview of the proposed \textit{ContactFlowFilter} framework.
        The proposed method samples particles from two complementary proposal distributions: 
        \textbf{Top}: The PF’s temporal prediction involves sampling from the previous time step, random position updates, and importance weighting (indicated by pink circles proportional to their weights).
        The same procedure applies to the particles around the second contact $\bm{F}_2$; only $\bm{F}_1$ is visualized for clarity.
        \textbf{Bottom}: The flow matching samples (indicated by pink and blue points) are transported along a probability path $(p\fidx{0} \rightarrow p\fidx{1})$ to represent the contact points.
        \textbf{Right}: These proposals are merged to integrate temporal context and multi-modality, forming the combined distribution for the subsequent recursive prediction.
	}
	\label{fig:mixture_overview}
    \vspace{-4mm}
\end{figure*}

While PF-based approaches are well-suited for addressing \textit{temporal dependency} through recursive belief updates, they often fail to handle \textit{multi-modality} due to particle starvation \cite{thrun2001robust, thrun2002probabilistic, koval2015pose}. 
This well-known issue of PFs arises when the particles fail to populate high-belief regions. 
In contrast, learning-based approaches excel at estimating \textit{multi-modal} distributions.
For instance, to capture multi-modality, probabilistic mapping \cite{liang2021contact} evaluates probabilities for each region of a discretized state space, and generative models \cite{cdm} learn the underlying data distribution.
These feed-forward approaches, however, are typically non-recursive and trained on temporally independent samples, thus lacking a native mechanism for leveraging \textit{temporal dependency}. 
As for \textit{contact event detection}, it remains a challenge for both categories of approaches.
Although PFs recursively track contact states, they rely on hard-to-tune thresholds to identify discrete transitions. 
Feed-forward learning-based methods struggle to model these discrete transitions, as their non-recursive nature inherently restricts temporal reasoning.

In this paper, we propose a novel framework called \textit{ContactFlowFilter} that tackles the aforementioned challenges: our approach is built upon a recursive structure of PFs for capturing \textit{temporal dependency}, while a generative model serves as an auxiliary proposal distribution to track \textit{multi-modal} beliefs and detect \textit{contact events}.

\subsection{Contributions of the Article}
\label{subsec:intro_contribution}
As depicted in Fig.~\ref{fig:mixture_overview}, \textit{ContactFlowFilter} updates the contact belief by drawing particles from two complementary proposal distributions: the PF's temporal prediction and the generative auxiliary proposal.
By preserving the PF's recursive structure, the proposed method effectively captures \textit{temporal dependency}, while the generative component complements the PF by representing \textit{multi-modal} distributions.
Building upon this hybrid framework, we propose a robust method for \textit{contact event detection} that augments conventional criteria by combining temporal context with generative estimates.

To formally implement the mixture of two proposal distributions, we adopt the dual-sampling concept \cite{thrun2001robust}.
We first formulate the PF motion model as Gaussian sampling on the robot surface, providing a basis for integrating the generative model and deriving its importance weights.

The generative component is built upon a novel data representation designed to capture the underlying distribution of contact states.
We utilize a fixed-size array of paired contact points to represent up to two simultaneous contacts, which allows the model to infer the spatial relationships between the points.
This is supported by our proposed observation representation, where spatially aligned robot kinematics data, JTSs data, and base F/T sensor data are mapped as a sequence along the kinematic chain. 
A transformer encoder then processes this sequence to extract cross-modal correlations.

To achieve real-time performance for high-fidelity temporal tracking, we employ flow matching \cite{liu2022flow, frans2024one} as the generative backbone, leveraging linearized probability paths for few-step inference rather than curved diffusion trajectories \cite{ho2020denoising, song2020score, luo2022understanding}.
We also implement a computationally efficient motion model using pre-processed mesh data \cite{10161173}.

In summary, our contributions are centered on a \textbf{main framework (C1)}, which is realized through key \textbf{technical contributions (C2, C3)}:
\begin{itemize}[leftmargin=*, nosep]
\item \textbf{C1)} 
A novel hybrid framework, \textit{ContactFlowFilter}, that synergistically combines a generative model and a particle filter to address the key challenges of the contact estimation problem (\textit{multi-modality}, \textit{temporal dependency}, and \textit{contact event detection}) while ensuring real-time performance.
(Validated in simulation (Section~\ref{subsec:exp_simulation}) and real-world experiments (Section~\ref{subsec:exp_real} and supplementary video).)

\item \textbf{C2)}
A spatially aligned data representation and a transformer-based architecture tailored to extract cross-modal correlations, enabling the flow matching model to effectively learn the \textit{multi-modal} distribution of contact points.
(Validated in Section~\ref{subsec:exp_selection_generative}.)

\item \textbf{C3)}
A robust method for \textit{contact event detection} that leverages a combination of distributions generated by the flow matching model and the PF.
(Validated in Section~\ref{subsec:exp_manage_pset}.)
\end{itemize}

\input{table/related_works_method_comparison}

We present thorough validation in both simulation and real-world settings.
Comparative experiments confirm that \textit{ContactFlowFilter} achieves superior performance compared to existing methods.
To assess the estimation accuracy, we employ localization error metrics along with the defined dissimilarity based on the approximated likelihood of the time-dependent belief.
In the real-world experiment, the proposed method achieved the lowest dissimilarity to the target distribution, with localization errors of 0.36~$\mathrm{cm}$ and 0.88~$\mathrm{cm}$ while maintaining average execution times of 0.69~$\mathrm{ms}$ and 2.69~$\mathrm{ms}$ for single-contact and dual-contact scenarios, respectively.

\subsection{Organization of the Article}
\label{subsec:intro_organization}
The rest of this article is organized as follows. 
Section~\ref{sec:related_works} reviews related research, and Section~\ref{sec:preliminaries} presents the preliminaries.
Section~\ref{sec:naive_pf} introduces the baseline PF framework for contact estimation, followed by a discussion of its limitations.
We then present the proposed method, \textit{ContactFlowFilter}, in Section~\ref{sec:mixture_alg} and detail its implementation in Section~\ref{sec:imple_detail}.
Section~\ref{sec:exp_val} demonstrates the experimental validation of the proposed method. 
Finally, Section~\ref{sec:conclusion} concludes this~article.

\section{Related Work}
\label{sec:related_works}
Characterizing the state of external contacts can be divided into four categories. 
The first two, \textbf{contact detection} (determining if a contact exists) and \textbf{contact isolation} (identifying the affected link), confirm the presence of a contact. 
The latter two provide more detailed information: \textbf{contact localization} focuses solely on the contact's position, while \textbf{contact estimation} aims to determine both location and force simultaneously. 
This section covers the literature on proprioception-based contact localization and estimation, and a comparative summary is provided in Table~\ref{tab:method_review}.
A primary approach for these problems utilizes external force/torque values, which originate exclusively from contacts rather than the robot's own dynamics (see Section~\ref{subsec:ext_values} for details).

\subsection{Model-Based Approaches}
The line-of-force (LoF) method is a widely adopted technique that geometrically localizes a contact and identifies the corresponding force from the external wrench \cite{buondonno2016combining, 8059840, iskandar2021collision, popov2022analytical}.
Because this geometric construction yields a single, deterministic solution based solely on the current observation, it is inherently limited to single-contact scenarios. 
Although \cite{iskandar2021collision} demonstrates dual-contact estimation on a robot with redundant sensing (i.e., JTSs and F/T sensors at the base and end-effector), the LoF method lacks temporal reasoning; the estimation is instead enabled by the dual F/T sensor setup, which prevents the spatial mixing of forces.
Intuitively, the base F/T sensor localizes a contact along the robot's main kinematic chain, while the end-effector F/T sensor identifies a second contact on any attached tool.

Another family of methods is based on the PF \cite{10342345, manuelli2016localizing, 10161173, zwiener2019armcl, zurlo2023collision, bimbo2019collision, popov2021multi, job2023multiple}. 
As a recursive approach, the PF naturally incorporates temporal information, allowing it to track contact states over time.
In this framework, a set of particles, representing contact points, is updated and resampled at each time step to minimize an error metric, defined via quadratic programming (QP).
However, a well-known issue with standard PF-based algorithms is particle starvation \cite{thrun2001robust, thrun2002probabilistic, koval2015pose}, where the greedy resampling process causes all particles to converge to a single high-belief region (even to a local minimum), which prevents the particle set from representing a multi-modal distribution.

To address the challenge of multi-modality (i.e., the singularity problem), several approaches have been proposed. 
For instance, the methods in \cite{pang2021identifying, popov2021real} employ a non-greedy strategy to narrow down contact point candidates, evaluating each using a QP-based error metric.
However, extending these approaches to scenarios with more than one contact point is not straightforward.

While such non-greedy schemes effectively handle multi-modality, both they and the LoF method rely on a static formulation that cannot capture the time-dependent characteristic of the contact estimation problem.
In contrast, the PF framework inherently supports temporal reasoning, making it a more promising foundation. 
Therefore, this work focuses on advancing the PF-based approach by overcoming its own limitations.

A key limitation of existing PF-based approaches for scenarios with more than a single contact \cite{manuelli2016localizing, 10161173, zwiener2019armcl, popov2021multi, job2023multiple} is their reliance on a hard-to-tune threshold for contact event detection.
While these methods track each contact point with a dedicated particle set, managing the number of sets depends on a highly sensitive threshold. 
Such dependency limits the adaptation to contact changes, leading to detection errors such as false positives or missed contacts.
To overcome this, we introduce a robust particle set management scheme that utilizes the output of flow matching, as detailed in Section~\ref{subsec:modified_manage_p_set}.

Additionally, PF-based methods are mainly distinguished by their strategies for representing and updating particles on the robot's surface. 
These include projecting particles represented in Cartesian space onto the robot surface \cite{manuelli2016localizing, zurlo2023collision}, parameterizing the surface with a single variable \cite{10342345, job2023multiple}, and using graph-based searches directly on the mesh \cite{10161173, zwiener2019armcl, popov2021multi}.
In Section~\ref{subsec:pf_motion_model}, we introduce a highly efficient particle update method that performs Gaussian sampling directly on the robot's surface by using pre-computed geodesic distances from the mesh data.

\subsection{Learning-Based Approaches}
Recently, several learning-based approaches have been proposed for contact localization \cite{liang2021contact, zwiener2018contact, popov2020transfer, cdm, fu2025unitac}, which vary in their formulations.
For example, the probabilistic mapping approach in \cite{liang2021contact} treats the robot's link surfaces as 2D images and estimates the contact probability for each region. 
Other formulations include binary classification over predefined points on the robot's surface \cite{zwiener2018contact, popov2020transfer}, or regression of the contact location, either as a 3D point in Cartesian space \cite{fu2025unitac} or as a single 1D parameter on the robot's surface \cite{popov2020transfer}.
However, a common limitation of these approaches is that they do not estimate the contact force and inherently lack a mechanism for temporal reasoning.

Among these learning-based methods, the work most relevant to ours is the diffusion model-based approach presented in \cite{cdm}. 
While it employs a recursive inference scheme similar to the PF, relying on temporally independent training data forces the model to assume that the first contact point remains fixed.
It also shares the common limitation of other learning-based approaches in its inability to estimate contact force.
In contrast, by integrating the output of a generative model into the PF framework, the proposed method performs both temporal reasoning and contact force estimation, while maintaining the representational power of the generative model.
By utilizing a flow matching model, our approach further achieves significantly faster inference (15.97 $\mathrm{ms}$$\,\rightarrow\,$2.69 $\mathrm{ms}$) while even demonstrating substantial improvements in performance.

\section{Preliminaries}
\label{sec:preliminaries}

\subsection{Particle Filter}
\label{subsec:pf_background}
The PF is a non-parametric version of the Bayes filter that estimates the posterior distribution of robot states conditioned on sensor data \cite{thrun2002probabilistic, thrun2001robust}.
In the Bayes filter approach, the target posterior that we aim to estimate is commonly referred to as the \textit{belief}:
\begin{align}
\label{eq:def_belief}
    bel(x_t) := p(x_t \vert o_{0:t}),
\end{align}
which represents the probability distribution over the state $x_t$, conditioned on all past observations $o_{0:t}$.
In our setup, it is assumed that the state evolution depends solely on the system dynamics; i.e., there is no control input for the state $x_t$.
Under the Markov assumption (i.e., if the current state is known, future observations are independent of past observations), the following recursive equations are obtained from \eqref{eq:def_belief}:
\begin{align}
\label{eq:bayes_alg_1}
    \overline{bel}(x_t) &= \int p(x_t \vert x_{t-1}) bel(x_{t-1})\dd{x_{t-1}}, \\
\label{eq:bayes_alg_2}
    bel(x_t) &= \eta p(o_t \vert x_t )\overline{bel}(x_t),
\end{align}
where $\overline{bel}(x_t)$ is the belief after the prediction step, and $\eta$ is a normalization constant.
Such a recursive filtering algorithm operates starting from $t=0$ and a common setup for the initial belief $bel(x_0)$ is a uniform distribution when the initial state is unknown.
A fundamental limitation of this recursive approach, however, is that each belief update requires the functional form of all posteriors to be specified.
Instead, non-parametric approaches, such as the PF, represent the posterior using a finite number of samples, which simplifies the implementation.

The key idea of the PF is to approximate the belief \eqref{eq:def_belief} using a set of particles $\mathcal{X}_t$:
\begin{align}
\label{eq:goal_pf}
    bel(x_t) \approx \mathcal{X}_t = \{ x_{t}\pidx{m}, w\pidx{m} \}_{m=1}^{M},
\end{align}
where $x_t\pidx{m}$ is the $m^{\text{th}}$ particle with its corresponding weight denoted as $w\pidx{m}$, and $M$ is the total number of particles in~$\mathcal{X}_t$.

\subsubsection{Sampling}
\label{subsubsec:sampling_background}
The sampling procedure includes the motion model and the measurement model of the PF, which correspond to \eqref{eq:bayes_alg_1} and \eqref{eq:bayes_alg_2}, respectively.
First, after applying the motion model, the particle set is distributed as the predicted belief; i.e., $\overline{bel}(x_t) \approx \bar{\mathcal{X}}_t$.
The particle $x_t\pidx{m} \in \bar{\mathcal{X}}_t$ is obtained via the following sampling procedure:
\begin{align}
\label{eq:prev_p_sampling}
    x_{t-1}\pidx{m} &\sim bel(x_{t-1}), \\
\label{eq:motion_model_background}
    x_{t}\pidx{m} &\sim p(x_t \vert x_{t-1}\pidx{m}).
\end{align}
The resulting pair of particles $\langle x_{t-1}\pidx{m}, x_{t}\pidx{m} \rangle$ is distributed as the following distribution called the \textit{proposal distribution}:
\begin{align}
\label{eq:def_proposal_dist}
    q(x_{t}) := p(x_t \vert x_{t-1}) bel(x_{t-1}).
\end{align}

The measurement model determines the perceptual likelihood of the particle to compensate for the gap between the \textit{proposal distribution} from which the particles are drawn, and the following \textit{target distribution}: 
\begin{align}
\label{eq:target_pdf}
    \eta p(o_t \vert x_{t}\pidx{m})p(x_t\pidx{m} \vert x_{t-1}\pidx{m})bel(x_{t-1}\pidx{m}),
\end{align}
which can be derived from \eqref{eq:bayes_alg_1} and \eqref{eq:bayes_alg_2} for $\langle x_{t-1}\pidx{m}, x_{t}\pidx{m} \rangle$.
The weight of the particle is obtained as a ratio of the \textit{target distribution} \eqref{eq:target_pdf} to the \textit{proposal distribution} \eqref{eq:def_proposal_dist}:
\begin{align}
\begin{split}
\label{eq:derivation_weight}
    w\pidx{m} &=  \frac{ \eta p(o_t \vert x_{t}\pidx{m})p(x_t\pidx{m} \vert x_{t-1}\pidx{m})bel(x_{t-1}\pidx{m})}{p(x_t\pidx{m} \vert x_{t-1}\pidx{m}) bel(x_{t-1}\pidx{m})} \\
    &= \eta p(o_t \vert x_t\pidx{m}).
\end{split}
\end{align}

\subsubsection{Dual-Sampling}
\label{subsubsec:dual_sampling_background}
In contrast to the standard sampling procedure, dual-sampling \cite{thrun2001robust} generates particles based solely on observations in a non-recursive manner. 
These particles, referred to as \textit{dual particles}, are drawn from the \textit{dual-proposal distribution}, which is defined as follows:
\begin{align}
\label{eq:dual_proposal_def}
    \bar{q}(x_t) := \frac{p(o_t \vert x_t)}{\pi(o_t)},
\end{align}
where $\pi(o_t) = \int p(o_t \vert x_t')\dd{x_t'}$.
Given a dual particle, $x_t\pidx{m} \sim \bar{q}(x_t)$, the pair of particles $\langle x_{t-1}\pidx{m}, x_{t}\pidx{m} \rangle$ is distributed as
\begin{align}
\label{eq:dual_proposal_pair}
    \frac{p(o_t \vert x_t)}{\pi(o_t)} bel(x_{t-1}).
\end{align}
As in the standard sampling procedure, the weight of a dual particle is defined as
\begin{align}
\begin{split}
\label{eq:dual_weight}
    w\pidx{m} &= \left( \frac{p(o_t \vert x_t\pidx{m})}{\pi(o_t)} bel(x_{t-1}\pidx{m}) \right)^{-1} \\
    &\quad \eta p(o_t \vert x_{t}\pidx{m}) p(x_t\pidx{m} \vert x_{t-1}\pidx{m}) bel(x_{t-1}\pidx{m}) \\
    &\propto p(x_t\pidx{m} \vert x_{t-1}\pidx{m}),
\end{split}
\end{align}
which takes the same form as the distribution of the motion model in \eqref{eq:motion_model_background}.
Note that while \eqref{eq:motion_model_background} represents the process of sampling a particle from a conditional distribution given a single particle, \eqref{eq:dual_weight} computes the probability given a pair of particles.

\subsubsection{Importance Resampling}
\label{subsubsec:background_imp_resampling}
After the sampling procedure, the weighted particles (the pair of $x_t\pidx{m}$ and $w\pidx{m}$) are distributed as the current belief $bel(x_t)$.
To ensure that the belief is represented solely by the particles (without relying on weights), the importance resampling process is commonly applied \cite{10161173}, which involves drawing a new particle set based on their weights.

\subsection{Flow Matching Model}
\label{subsec:ddpm}
Generative models, such as diffusion models \cite{ho2020denoising, song2020score, luo2022understanding} and flow matching models \cite{frans2024one, liu2022flow}, are designed to generate a new sample $\hat{x}$ that follows the data distribution $p(x)$.
In practice, this distribution is accessible only empirically.

In the context of flow matching, $p\fidx{k}$ denotes the probability path that connects a simple distribution to the data distribution.
At the boundary, $p\fidx{0}$ is typically defined as the standard Gaussian $\mathcal{N}(0, I)$, while $p\fidx{1}$ approximately corresponds to the target data distribution $p$.
By adopting the optimal transport flow matching objective \cite{liu2022flow}, $x\fidx{k}\sim p\fidx{k}(x)$ is defined as a linear interpolation between a data point $x\fidx{1} \sim \mathcal{D}$ and a noise point $x\fidx{0} \sim p\fidx{0}(x)$:
\begin{align}
\label{eq:interpolant}
x\fidx{k} &= (1-k)x\fidx{0} + k x\fidx{1}, \\
v\fidx{k} &= x\fidx{1} -x\fidx{0},
\end{align}
where $v\fidx{k}$ denotes the vector field associated with the probability path $p\fidx{k}$.
For a sampled pair $\langle x\fidx{0},x\fidx{1} \rangle$, the flow matching objective is defined as 
\begin{align}
\label{eq:fm_objective}
{\mathcal{L}}_\theta = \displaystyle \mathop{\mathbb{E}}_{k,x\fidx{0},x\fidx{1}} \norm{{v\fidx{k}} - {v}_{\theta}(x\fidx{k}, k)}^2,
\end{align}
where $\theta$ represents learnable network parameters, and $k \sim \mathcal{U}(0,1)$, which is the uniform distribution.

A sample is generated by starting from a noise input $x\fidx{0} \sim \mathcal{N}(0, I)$ and solving the ordinary differential equation (ODE) defined by the learned vector field $v_{\theta}$, which transports the sample along the probability path from $p\fidx{0}$ to $p\fidx{1}$.

\subsection{Contact Induced Sensor Measurements}
\label{subsec:ext_values}
Our proprioception setup includes the JTSs and the base F/T sensor (see Fig.~\ref{fig:intro_figure}).
Suppose that $n_c \in \{1,2\}$ contacts exist on the $n_q$-DOF robotic arm (up to one contact per link), and let the location and force of the $i^{\text{th}}$ contact be denoted as $\bm{r}_i \in \mathbb{R}^3$ and $\bm{F}_i \in \mathbb{R}^3$, respectively.
Then, the relations between the contacts and the available proprioception are given by:
\begin{align}
\label{eq:external_torque}
    \bm{\tau}_{ext} &= \sum_{i=1}^{n_c} \bm{J}_{i}^{T}(\bm{q}, \bm{r}_{i}) \bm{F}_{i}, \\
\label{eq:external_base_wrench}
    \bm{\mathcal{F}}_{ext} &= \sum_{i=1}^{n_c} \underbrace{\begin{bmatrix} \mathbf{I}_{3\times3}  \\ [\bm{r}_{i}]_{\times} \end{bmatrix}}_{\triangleq \bm{A}_{i}(\bm{r}_{i})} \bm{F}_{i},
\end{align}
where $\bm{\tau}_{ext} \in \mathbb{R}^{n_q}$ is the external joint torque,
$\bm{\mathcal{F}}_{ext} \in \mathbb{R}^6$ is the base wrench due to contacts,
$\bm{q}\in\mathbb{R}^{n_q}$ denotes the joint variables, 
$\bm{J}_i(\bm{q},\bm{r}_i)$ is the associated positional Jacobian matrix at the contact point, 
and $[\cdot]_{\times}$ denotes the skew-symmetric matrix such that $[\bm{a}]_{\times}\bm{b} = \bm{a} \times \bm{b}$ for any $\bm{a}, \bm{b} \in \mathbb{R}^3$.
For brevity, we define the vector $\bm{\mathcal{W}}_{ext}=(\bm{\tau}_{ext}, \bm{\mathcal{F}}_{ext})$.

\subsubsection{External Joint Torque Estimation}
\label{subsubsec:tau_ext}
We consider the following link dynamics of a flexible joint robot model:
\begin{align}
\label{eq:robot_dyn}
    \bm{M}(\bm{q})\ddot{\bm{q}} + \bm{C}(\bm{q}, \dot{\bm{q}}) \dot{\bm{q}} + \bm{g}(\bm{q}) = \bm{\tau}_j + \bm{\tau}_{ext},
\end{align}
where $\bm{M}(\bm{q})$ is the link-side inertia matrix, 
$\bm{C}(\bm{q},\dot{\bm{q}})$ is the Coriolis/centrifugal matrix that satisfies $\dot{\bm{M}} = \bm{C}+\bm{C}^T$, 
$\bm{g}(\bm{q})$ is the gravity vector, 
and $\bm{\tau}_{j}$ is the joint torque measured by the JTSs.
The estimated external joint torque, denoted as $\hat{\bm{\tau}}_{ext}$, can be obtained using the following momentum-based disturbance observer \cite{de2006collision,kim2015design}:
\begin{align}
\label{eq:DOB}
    \hat{\bm {\tau}}_{ext}(t) &= \mathbf{K}_{\text{o}}\left\{\bm{p}-\int _0^t 	\left(\bm{\tau }_{j} + \bm{n}(\bm{q}, \dot{\bm{q}}) +\hat{\bm{\tau}}_{ext}\right) \dd{s}\right\},
\end{align}
where $\bm{p} = \bm{M}(\bm{q})\dot{\bm{q}}$, 
$\bm{n}(\bm{q}, \dot{\bm{q}}) = \bm{C}^T(\bm{q},\dot{\bm{q}})\dot{\bm{q}} - \bm{g}(\bm{q})$,
and $\mathbf{K}_{\text{o}}$ is a diagonal observer gain matrix.
The observer output can be regarded as a first-order low-pass filtered estimate of the actual $\bm{\tau}_{ext}$ \cite{de2006collision, haddadin2008collision}.

\subsubsection{External Base Wrench Estimation}
\label{subsubsec:ext_base_wrench}
We employ the method proposed in \cite{buondonno2016combining} to explicitly compensate for the dynamic effects of the robot itself from the base F/T sensor measurements $\bm{\mathcal{F}}_{b}$.
First, we calculate the nominal joint acceleration $\ddot{\bm{q}}_n$ by subtracting the estimated external joint torque \eqref{eq:DOB} from the robot dynamics \eqref{eq:robot_dyn}.
The nominal acceleration corresponds to a contact-free scenario where the nominal robot has the same joint configuration as the real robot, but without any external contact forces.
For the nominal robot, the interaction wrench between the base F/T sensor and the robot base link represents the dynamic effects that we aim to compensate for.
This wrench is efficiently computed using the recursive Newton-Euler algorithm (RNEA) as $\bm{\mathcal{F}}_n = \text{RNEA}(\bm{q}, \dot{\bm{q}}, \ddot{\bm{q}}_n)$, where $\bm{\mathcal{F}}_n$ is the nominal base wrench exerted by the base F/T sensor on the robot base link.
The estimate $\hat{\bm{\mathcal{F}}}_{ext}$ is then obtained by subtracting the nominal wrench from the measurements: $\hat{\bm{\mathcal{F}}}_{ext} = \bm{\mathcal{F}}_{b} - \bm{\mathcal{F}}_{n}$.

\section{Particle Filter for Contact Estimation}
\label{sec:naive_pf}
This section presents the baseline PF algorithm for contact estimation and identifies its limitations, motivating the subsequent section.
Throughout this paper, we use the following notation:
\begin{itemize}[leftmargin=*, nosep]
\item[-] $\mathcal{X}_{t,i}$: the $i^{\text{th}}$ set of particles at time $t$
\item[-] $\bm{x}_{t,i}\pidx{m}$: the $m^{\text{th}}$ particle in $\mathcal{X}_{t,i}$
\item[-] $w_i\pidx{m}$: the weight of the particle $\bm{x}_{t,i}\pidx{m}$
\item[-] $\mathfrak{X}_t$: the collection of particle sets, i.e., {\small $\mathfrak{X}_t = \{\mathcal{X}_{t,1},\cdots , \mathcal{X}_{t,n_c}\}$}
\end{itemize}

\subsection{Problem Statement}
\label{subsec:problem_statements}
Let $\mathbf{r}_t = \{\bm{r}_{t,i}\}_{i=1}^{n_c}$ denote the set of contact points at time $t$.
The \textit{contact localization} problem is formulated as estimating the following \textit{belief}:
\begin{align}
\label{eq:target_posterior}
    bel(\mathbf{r}_t) := p(\mathbf{r}_t \vert \bm{O}_{0:t}),
\end{align}
which is the probability distribution over the contact points $\mathbf{r}_t$, conditioned on all past observations $\bm{O}_{0:t}$.
The observation at each time step $\bm{O}_t$ consists of (i) kinematic quantities (e.g., link frame pose and Jacobian), and (ii) the estimated external value $\hat{\bm{\mathcal{W}}}_{ext}$. 
These components will be discussed in more detail later.
The goal of the PF algorithm is to approximate the belief \eqref{eq:target_posterior} using a collection of particle sets, i.e., $ bel(\mathbf{r}_t) \approx \mathfrak{X}_t$, where $\mathcal{X}_{t,i} \in \mathfrak{X}_t$ estimates the $i^{\text{th}}$ contact point $\bm{r}_{t,i}$.
The \textit{contact force identification} problem is simultaneously addressed within the PF framework via the measurement model (detailed in Section~\ref{subsec:pf_measurement_model}).

Note that we place no restrictions on robot configurations or contact locations; by utilizing the base F/T sensor, we retain observability even when the Jacobian $\eqref{eq:external_torque}$ is rank-deficient, such as during kinematic singularities or contacts on proximal links (e.g., Link 1).

\subsection{Motion Model}
\label{subsec:pf_motion_model}
The prediction step of the PF begins by drawing particles from the belief at the previous time step, i.e., $\bm{x}_{t-1,i}^{(m)} \sim bel(\mathbf{r}_{t-1})$.
The particle for the current time step $\bm{x}_{t,i}\pidx{m}$ is then sampled from the distribution of the motion model, $p(\bm{x}_t \vert \bm{x}_{t-1,i}\pidx{m})$.
In the absence of control input for the state evolution, a common choice is to use a random walk \cite{10161173, manuelli2016localizing, zwiener2019armcl} by adding random perturbations.
To express this mathematically, we represent the motion model as a Gaussian:
\begin{align}
\label{eq:motion_model_gaussian_pdf}
    p(\bm{x}_t \vert \bm{x}_{t-1,i}\pidx{m}) := \mathcal{N}(\bm{\mu}, \bm{\Sigma}),
\end{align}
where $\bm{\mu}= \bm{x}_{t-1,i}\pidx{m}$ and $\bm{\Sigma}=\sigma^2\mathbf{I}$ represent the mean and the covariance matrix of the Gaussian distribution, respectively, with $\sigma$ denoting the standard deviation.
However, particles sampled from \eqref{eq:motion_model_gaussian_pdf} may not lie on the robot surface, requiring additional, time-consuming processing (e.g., projection onto the surface).
To address this, we propose a highly efficient method that enables sampling directly on the robot surface.

\begin{figure}[t!]
\centering
    {\includegraphics[width=\columnwidth, trim = {0 0px 0 0px}, clip]{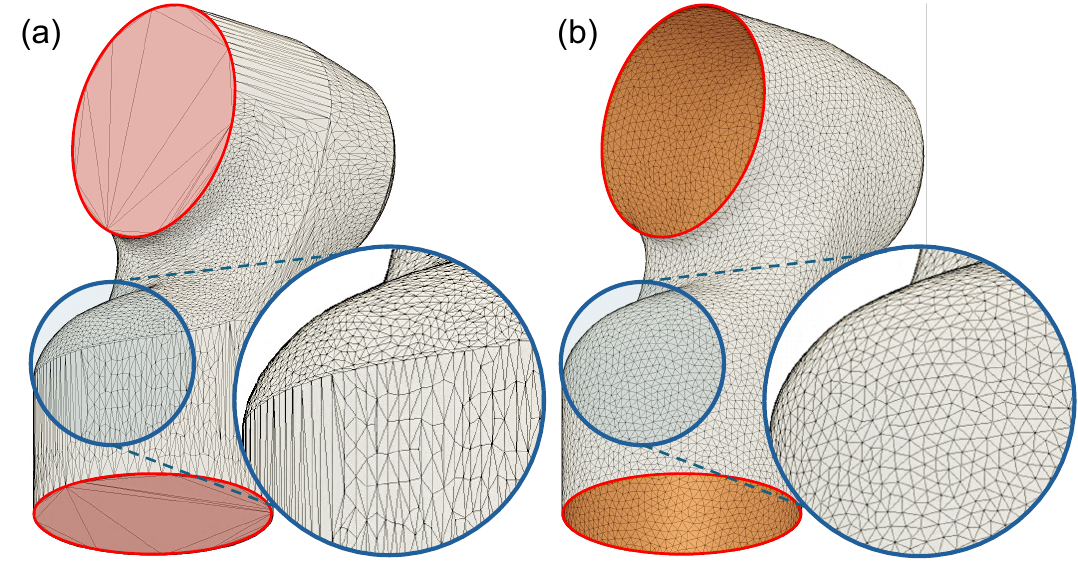}}
    \vspace{-6mm}
    \caption{
    The Link 1 meshes of the Franka Emika Panda robot are used for (a) visualization and (b) motion model computation.
    Non-contact faces are removed (red circle), and the mesh is refined into uniform triangular faces (blue circle).
    }
    \label{fig:mesh_preprocessing}

\end{figure}

\subsubsection{Mesh Preprocessing}
Adopting the particle representation from \cite{10161173}, we utilize the robot's mesh data to represent particle positions, ensuring that particles remain constrained to the robot surface.
The position of the particle is given by a link index and a face index; i.e., $\bm{x}_{t,i}\pidx{m} = (\texttt{link\_idx}, \texttt{face\_idx})$.
To this end, we perform mesh preprocessing as shown in Fig.~\ref{fig:mesh_preprocessing}.
Using Blender \cite{Hess:2010:BFE:1893021}, we manually remove the faces where contact cannot occur, and apply an isotropic remeshing algorithm \cite{yan2009isotropic} to ensure that the faces are of uniform size.
In our implementation, the target edge length of the mesh faces is set to approximately 0.36~$\mathrm{cm}$.

\begin{figure}[t!]
\centering
    {\includegraphics[width=\columnwidth, trim = {0 0px 0 0px}, clip]{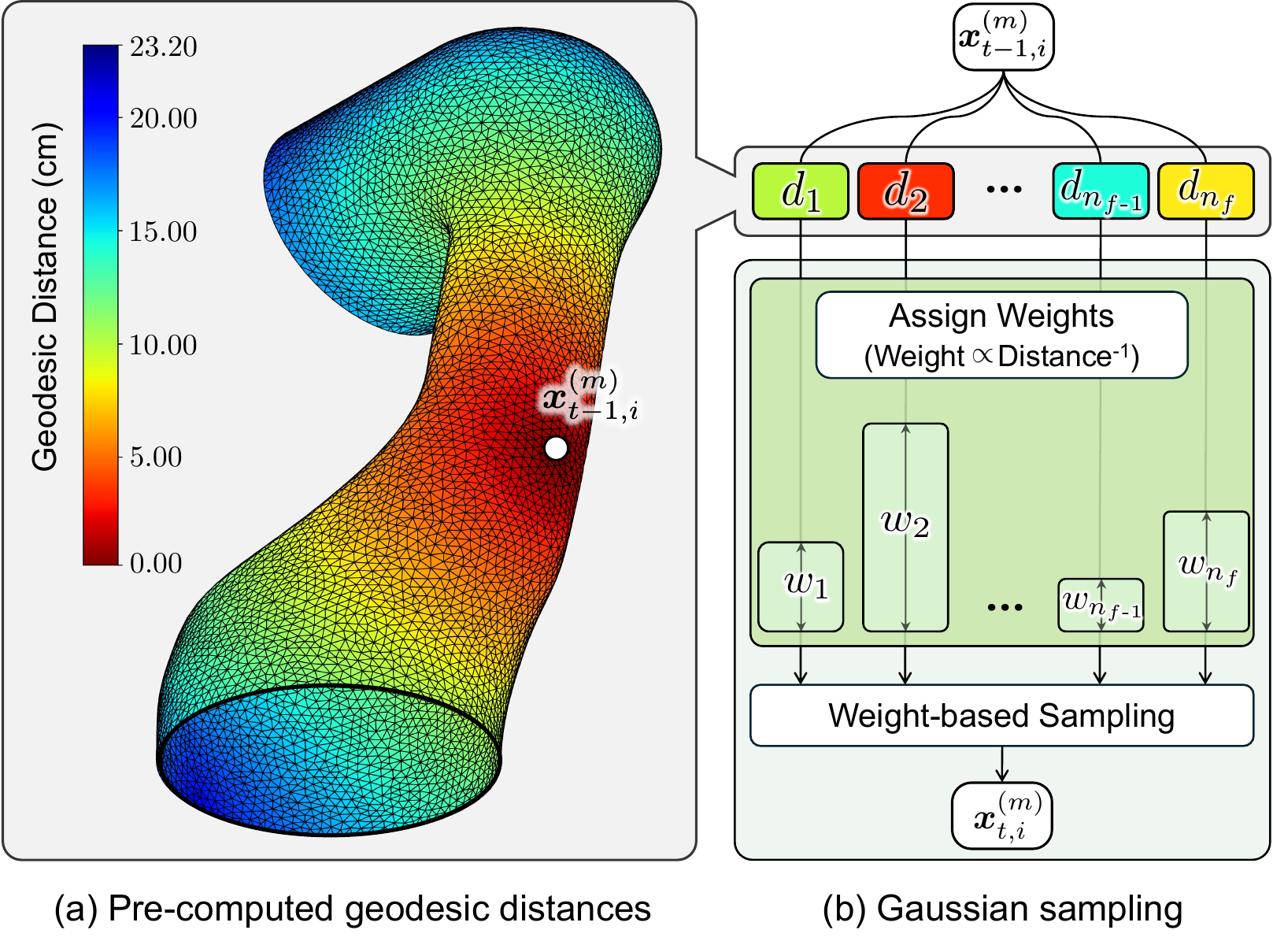}}
    \vspace{-8.1mm}
    \caption{
    The proposed motion model consists of two steps:
    (a) For the given particle $\bm{x}_{t-1,i}\pidx{m}$, the geodesic distances to all other faces $\{d_j\}_{j=1}^{n_f}$ are retrieved from a pre-computed lookup table.
    (b) The weights $\{w_j\}_{j=1}^{n_f}$ are set inversely proportional to the geodesic distances; the particle for current time step $\bm{x}_{t,i}\pidx{m}$ is then sampled from the resulting distribution.
    }
    \label{fig:motion_model}
    \vspace{-2mm}
\end{figure}

\subsubsection{Approximating the Gaussian on the Surface}
The proposed motion model is illustrated in Fig. \ref{fig:motion_model}.
We pre-compute the geodesic distances for all face pairs on the robot's surface and store these values in a lookup table (see Fig.~\ref{fig:motion_model}(a)).
This enables efficient retrieval of geodesic distances from a source face to all other faces.
Using this lookup table, \eqref{eq:motion_model_gaussian_pdf} is approximated by a discretized distribution (see Fig.~\ref{fig:motion_model}(b)) with weights proportional to $\frac{1}{\sigma\sqrt{2\pi}} \exp\left({-0.5(d/\sigma)^2}\right)$,
where $d \in \mathbb{R}^+$ is the geodesic distance retrieved from the table.

\begin{rem}
Since computationally intensive tasks, such as mesh preprocessing and geodesic distance computation, are performed offline, the online execution of the proposed motion model is highly efficient.
Specifically, our implementation requires approximately 0.6~$\mathrm{\mu s}$ to update a single particle.
\end{rem}

\subsection{Measurement Model}
\label{subsec:pf_measurement_model}
The measurement model determines how well the given particle explains the observation by computing $p(\bm{O}_t \vert \bm{x}_{t,i}\pidx{m})$.
We first formulate the \textit{contact force identification problem} as a QP and use its solution to define the weight of a particle.

\begin{figure}[t!]
\centering
    {\includegraphics[width=\columnwidth, trim = {0px 0px 0 0px}, clip]{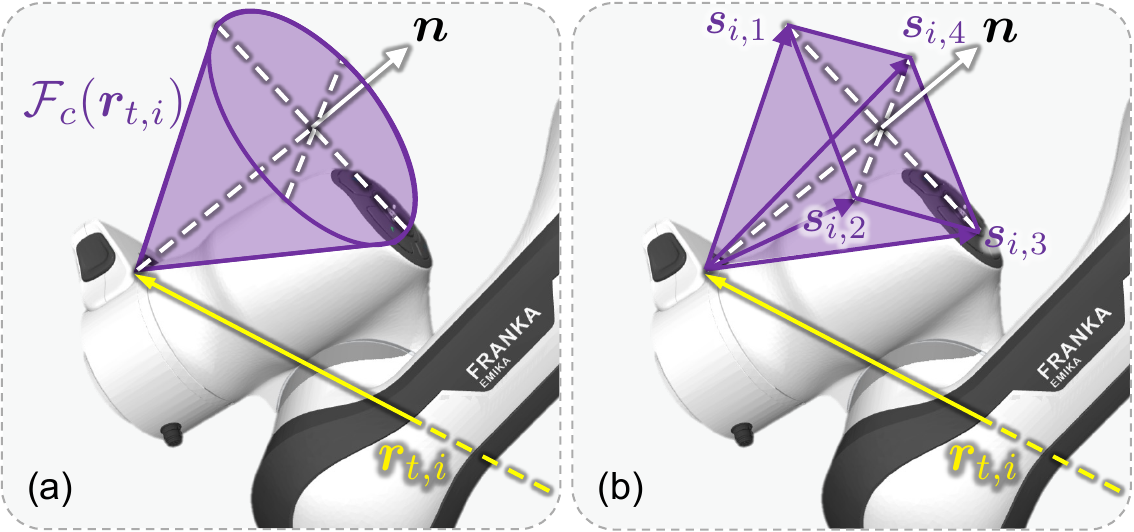}}
    \vspace{-6mm}
    \caption{
    (a) For each contact point $\bm{r}_{t,i}$, the friction cone $\mathcal{F}_c$ is defined by the surface normal $\bm{n}$.
    (b) The friction cone is approximated as a polyhedral cone spanned by support vectors $\{\bm{s}_{i,j}\}_{j=1}^{4}$.
    }
    \vspace{-3mm}
    \label{fig:friction_cone}
\end{figure}

\subsubsection{QP Formulation}
Let $\mathbf{F}_t$ denote the set of contact forces at time $t$; i.e., $\mathbf{F}_t = \{ \bm{F}_{t,i}\} ^{n_c} _{i=1}$.
Using \eqref{eq:external_torque} and \eqref{eq:external_base_wrench}, we formulate the contact force identification problem as an optimization problem:
\begin{equation}
\centering
\begin{gathered}
\label{eq:qp_before}
    \min_{\mathbf{F}_t} \left \| {\hat{\bm{\mathcal{W}}}_{ext} - \sum_{i=1}^{n_c}
    \begin{bmatrix}
        \bm{J}_{i}(\bm{q},\bm{r}_{t,i}) & \bm{A}_i(\bm{r}_{t,i})^T
    \end{bmatrix}^T \bm{F}_{t,i}
    } \right \| ^2 \\ 
\text{subject to } \bm{F}_{t,i} \in \mathcal{F}_c(\bm{r}_{t,i}),\, \forall i \in \{ 1, \cdots, n_c\},
\end{gathered}
\end{equation}
where $\mathcal{F}_c(\bm{r}_{t,i})$ is the friction cone at the $i^{\text{th}}$ contact point $\bm{r}_{t,i}$.
Here, the friction cone constraint accounts for the point contact assumption, where only pushing forces (i.e., no pulling forces) can be applied to the robot surface (see Fig. \ref{fig:friction_cone}(a)). \footnote{In practice, tuning the friction cone balances physical strictness and algorithmic convergence: an expanded force space smooths the QP error across the surface, yielding a continuous update path for particles.}
Note that the decision variables in \eqref{eq:qp_before} are the contact forces, while the contact points are treated as inputs.
In other words, \eqref{eq:qp_before} identifies the contact forces that best explain the observation, given the contact point locations.

To make the friction cone constraint tractable, $\mathcal{F}_c(\cdot)$ is approximated as a polyhedral cone as shown in Fig. \ref{fig:friction_cone}(b).
The $i^{\text{th}}$ contact force is expressed as $\bm{F}_{t,i} = \bm{S}_{i}\bm{f}_i$,
where $\bm{S}_{i} = [\bm{s}_{i,1}, \dots, \bm{s}_{i,4}]$ is the matrix of support vectors ($\bm{s}_{i,\bullet} \in \mathbb{R}^3$) spanning the polyhedral friction cone, and $\bm{f}_i = (f_{i,1}, \dots, f_{i,4})$ is the vector of corresponding weights ($f_{i,\bullet} \leq 0$).
Substituting this force expression into \eqref{eq:qp_before}, we reformulate the optimization problem as a QP:
\begin{equation}
\centering
\begin{gathered}
\label{eq:qp_after}
    QP(\bm{O}_t \vert \mathbf{r}_{t}) = 
    \min_{\mathbf{f}_{t}} \left \| {\hat{\bm{\mathcal{W}}}_{ext} - \mathbf{Q}^T\mathbf{f}_t
    } \right \| ^2 \\ 
\text{subject to } \bm{f}_i \preceq 0, \,\, \forall i \in \{ 1, \cdots, n_c\},
\end{gathered}
\end{equation}
where $\mathbf{f}_{t} = (\bm{f}_{1}, \cdots, \bm{f}_{n_c})\in \mathbb{R}^{4n_c}$,
and $\mathbf{Q}\in \mathbb{R}^{4n_c\times(n_q+6)}$ is a matrix constructed from $\bm{S}_i^T\bm{J}_i$ and $\bm{A}_{i}\bm{S}_i$ matrices.
Intuitively, a particle with a smaller QP error corresponds to a more probable contact point.

\subsubsection{Importance Factor}
\label{subsubsec:weight_pf}
Consider a single-contact scenario ($n_c=1$) where the number of particle sets is properly adjusted (as presented in Section~\ref{subsec:pf_manage_p_set}).
In this case, the weight of a particle in $\mathcal{X}_{t,1}$ is defined using the QP error metric \eqref{eq:qp_after} as
\begin{align}
\begin{split}
w_{1}\pidx{m} &=  p(\bm{O}_t \vert \bm{x}_{t,1}\pidx{m})  \\ 
&:= \eta \exp \left( -\alpha QP(\bm{O}_t \vert \{ \bm{x}_{t,1}\pidx{m} \}) \right),
\end{split}
\end{align}
where $\alpha \in \mathbb{R}^+$ controls the steepness of the exponential function, thereby determining the sensitivity to the QP error.

However, in the case of $n_c$ contacts, the particle sets for all contacts must be considered jointly, as the input to the QP \eqref{eq:qp_after} consists of contact point candidates.
Evaluating all possible combinations of particles poses a significant computational burden. 
For instance, in a dual-contact scenario with $M=100$ particles per set, the total number of QP computations would be $100^2$.
In order to reduce such a burden, the following assumption is introduced:
\begin{assumption}
\label{ass:measurement_model}
When updating the weight of a particle in $\mathcal{X}_{t,i}$, the other particle sets $\mathfrak{X}_t \setminus\mathcal{X}_{t,i}\!=\!\{\mathcal{X}_j \vert \forall j\!=\!1,\!\cdots\!,n_c, j\!\neq\! i \}$ have already converged to the true contact points.
\end{assumption}
This assumption allows us to jointly consider only a single particle from each set in $\mathfrak{X}_t \setminus\mathcal{X}_{t,i}$; i.e., a particle set is represented by a single particle.
To introduce randomness, we sample one representative particle according to its weight, rather than selecting the particle with the highest weight.
The weight of a particle in $\mathcal{X}_{t,i}$ is defined as
\begin{align}
\begin{split}
\label{eq:weight_multi}
w_{i}\pidx{m} &=  p(\bm{O}_t \vert \bm{x}_{t,i}\pidx{m}) \\ 
&:= \eta \exp \left( -\alpha QP(\bm{O}_t \vert \{ \bm{x}_{t,i}\pidx{m}, \mathbf{x}^{\#}(\mathfrak{X}_t \setminus\mathcal{X}_{t,i}) \}) \right),
\end{split}
\end{align}
where $\mathbf{x}^{\#}(\mathfrak{X}_t \setminus\mathcal{X}_{t,i}) =\{\bm{x}^{\#}(\mathcal{X}_j)\vert \forall \mathcal{X}_j \in \mathfrak{X}_t \setminus\mathcal{X}_{t,i} \}$, and $\bm{x}^{\#}(\mathcal{X}_j)$ is a particle sampled from $\mathcal{X}_j$ based on the weights.
\begin{center}
\begin{algorithm}[htb!]
	\caption{Manage-Particle-Set ($\mathfrak{X}_t$)}
    \begin{algorithmic}[1]
    \preComment{increasing condition}
	\If {$\epsilon (\bm{O}_t,\mathfrak{X}_t) > \bar{\epsilon}$}
	    \State add $\mathcal{X}_{init} $ to $\mathfrak{X}_t$
	    \State \Return{$\mathfrak{X}_t$}
	\EndIf
    \preComment{decreasing condition}
	\For {$\mathcal{X}_{t,i}$ in $\mathfrak{X}_t$}
	    \If{$\epsilon (\bm{O}_t,\mathfrak{X}_t\setminus \mathcal{X}_{t,i} ) < \bar{\epsilon}$}
	        \State discard $\mathcal{X}_{t,i}$ from $\mathfrak{X}_t$
	        \State \Return{$\mathfrak{X}_t$}
	    \EndIf
	\EndFor
	\State \Return{$\mathfrak{X}_t$}
	\end{algorithmic}
	\label{alg:manage_pf}
\end{algorithm}
\end{center}
\subsection{Particle Set Management}
\label{subsec:pf_manage_p_set}
The number of particle sets (the size of $\mathfrak{X}_t$) must be properly adjusted to match the number of active contacts.
In this subsection, we present the method of \cite{manuelli2016localizing} as a preliminary, which will be extended further in Section~\ref{subsec:modified_manage_p_set}.

In Algorithm~\ref{alg:manage_pf}, the error metric is defined as $\epsilon (\bm{O}_t,\mathfrak{X}_t) = QP(\bm{O}_t \vert \mathbf{x}^{*}(\mathfrak{X}_t))$, where $\mathbf{x}^{*}(\mathfrak{X}_t) = \{ \bm{x}^*(\mathcal{X}_{t,i}) \vert \forall \mathcal{X}_{t,i} \in \mathfrak{X}_t\}$, and $\bm{x}^*(\mathcal{X}_{t,i})$ is the particle with the highest weight in $\mathcal{X}_{t,i}$.
If $\mathfrak{X}_t = \emptyset$, $\epsilon (\bm{O}_t,\mathfrak{X}_t)$ is set to $\| \hat{\bm{\mathcal{W}}}_{ext} \|^2$ to initialize the PF algorithm.
In lines 1\,–\,4, if the QP error of the current collection $\mathfrak{X}_t$ exceeds the threshold $\bar{\epsilon}$, indicating that the current observation cannot be explained by $\mathfrak{X}_t$, then a new particle set $\mathcal{X}_{init}$ is added.
Here, the particles in $\mathcal{X}_{init}$ are uniformly distributed over the robot surface.
Similarly, in lines 5\,–\,10, we check whether the current observation can be sufficiently explained without a certain particle set $\mathcal{X}_{t,i}$. 
If that is the case, $\mathcal{X}_{t,i}$ is removed from $\mathfrak{X}_t$.

\begin{algorithm}
	\caption{CPF ($\mathfrak{X}_{t-1}$)}
	\begin{algorithmic}[1]
	\State $\bar{\mathfrak{X}}_t \leftarrow$ Motion-Model$(\mathfrak{X}_{t-1})$ \codeComment{in Sec.~\ref{subsec:pf_motion_model}}
	\State ${\mathfrak{X}}_t \leftarrow$ Measurement-Model$(\bar{\mathfrak{X}}_{t})$ \codeComment{in Sec. \ref{subsec:pf_measurement_model}}
	\State $\mathfrak{X}_t \leftarrow$ Importance-Resampling$({\mathfrak{X}}_t)$ \codeComment{in Sec.~\ref{subsubsec:background_imp_resampling}}
	\State $\mathfrak{X}_t \leftarrow$ Manage-Particle-Set$(\mathfrak{X}_t)$ \codeComment{in Sec.~\ref{subsec:pf_manage_p_set}}
	\State \textbf{Return} $\mathfrak{X}_t$
	\end{algorithmic}
\label{alg:cpf_alg}
\end{algorithm}
\subsection{Contact Particle Filter}
\label{subsec:pf_cpf}
Algorithm \ref{alg:cpf_alg} outlines the Contact Particle Filter (CPF), which sequentially executes the procedures detailed in the previous sections.
In line 3, we employ the resampling process described in Section~\ref{subsubsec:background_imp_resampling}.
Through this recursive algorithm, the particles in $\mathfrak{X}_t$ approximate the belief \eqref{eq:target_posterior}, where the $i^{\text{th}}$ particle set ($\mathcal{X}_{t,i}\in \mathfrak{X}_t$) corresponds to the $i^{\text{th}}$ contact point.
The estimated $i^{\text{th}}$ contact force is derived from the corresponding QP solution \eqref{eq:qp_after} computed for the weight update of the $i^{\text{th}}$ particle set $\mathcal{X}_{t,i}$.

\subsection{Limitation of CPF}
\label{subsec:cpf_limitation}
\begin{figure}[t!]
\centering
    {\includegraphics[width=\columnwidth, trim = {0 0px 0 0px}, clip]{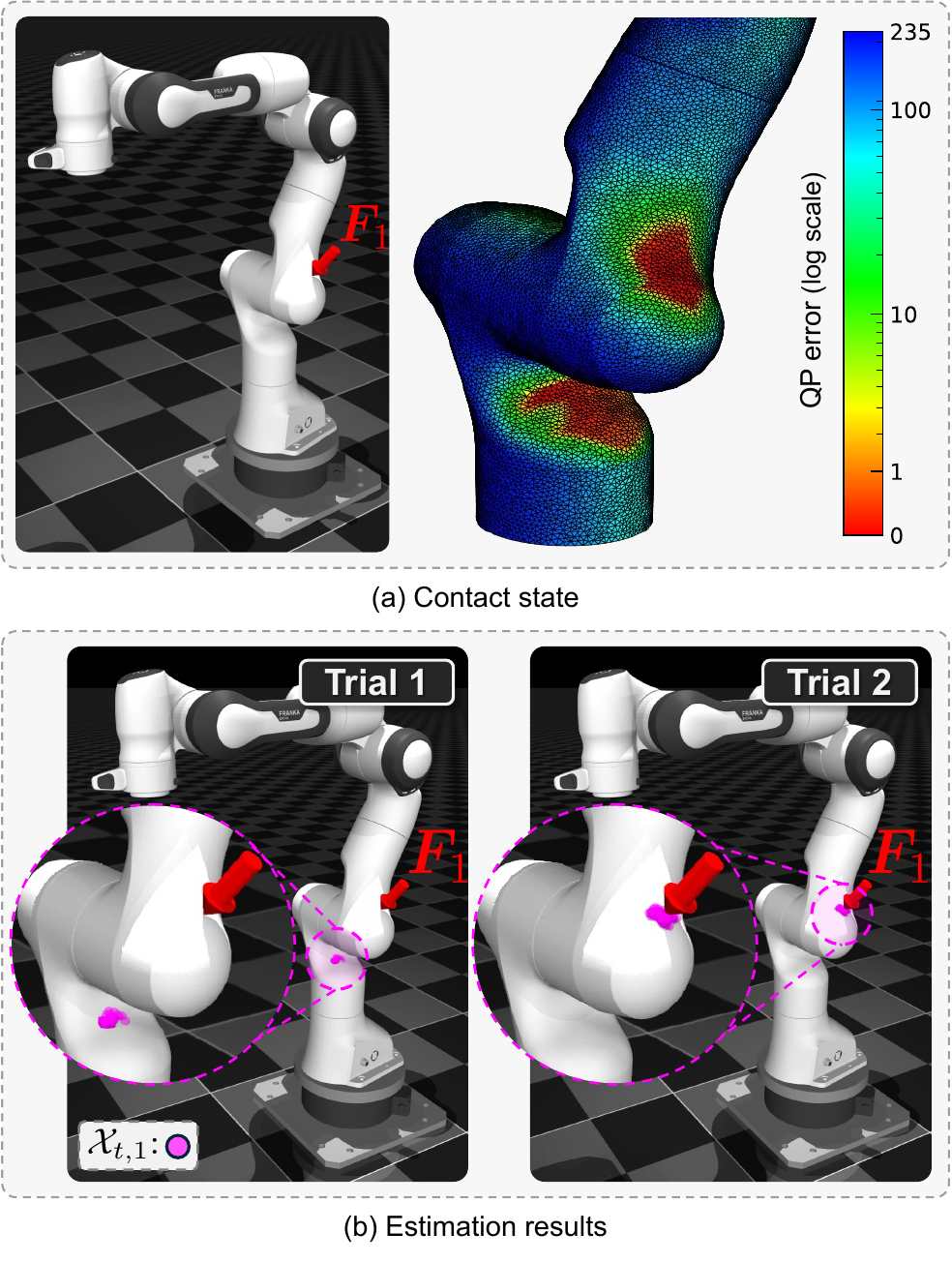}}
    \vspace{-9mm}
    \caption{
    (a) The results of the QP defined in \eqref{eq:qp_after} are visualized over the entire surface for a single contact ($\bm{F}_1$) on link 2,
     where high-belief (low QP error) regions are observed on both links 1 and 2.
    (b) The conventional PF algorithm populates the particle set $\mathcal{X}_{t,1}$ on only a single link in each execution (see Trial 1 and Trial 2).
    } 
    \label{fig:multi_modal_qp_map}
\end{figure}

\subsubsection{Particle Starvation}
\label{subsubsec:pf_starvation}
A fundamental challenge in PFs is the failure of particles to populate high-belief regions.
This particle starvation \cite{thrun2001robust, thrun2002probabilistic, koval2015pose} primarily stems from the mismatch between the \textit{proposal} and \textit{target} distributions, as captured in \eqref{eq:derivation_weight}. 
The motion model is a simple random walk that does not guide particles toward high-belief regions.
Convergence relies solely on the importance resampling process, requiring a high $\alpha$ value in \eqref{eq:weight_multi} to ensure that only particles with higher weights survive in a more greedy manner.
However, coupled with the intrinsic sparsity of the solution space, this setting further constricts $p(\bm{O}_t \vert \cdot)$, resulting in isolated, narrow high-belief regions surrounded by low-probability areas (e.g., see Fig.~\ref{fig:multi_modal_qp_map}(a)).
This prevents particles from traversing to other modes, causing the PF to become trapped in local minima or fail to capture the multi-modality. 
For example, as depicted in Fig.~\ref{fig:multi_modal_qp_map}(b), in a scenario with two belief modes, the CPF algorithm converges to only a single mode, unable to properly represent the distribution.

Although the proposed motion model is computationally efficient, it has a key limitation: particles cannot transition across links, as the geodesic distances are not defined across different links in the pre-computed lookup table. 
If no particles are present on the actual contact link, they cannot move to the high-belief regions.

\begin{figure}[t!]
\centering
    {\includegraphics[width=\columnwidth, trim = {0 0px 0 0px}, clip]{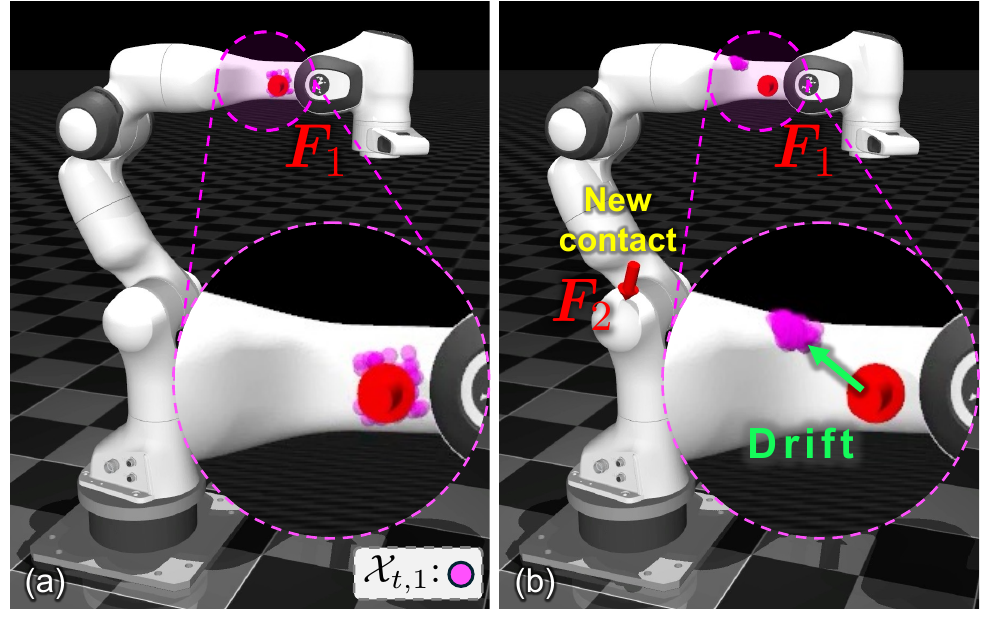}}
    \vspace{-8mm}
    \caption{
    (a) The first particle set $\mathcal{X}_{t,1}$ converges to the first contact point.
    (b) After the second contact ($\bm{F}_2$) occurs, the delayed initialization of $\mathcal{X}_{t,2}$ causes the existing set $\mathcal{X}_{t,1}$ to drift away from the first contact point.
    }
    \label{fig:particle_drift}
    
\end{figure}

\subsubsection{Particle Drift}
\label{subsubsec:pf_drift_v2}
When the number of contacts $n_c$ changes, particle drift occurs as the low-pass filtered estimate $\hat{\bm{\mathcal{W}}}_{ext}$ gradually adapts to such discrete transitions in contact states.
As illustrated in Fig.~\ref{fig:particle_drift}, this may result in particles erroneously drifting to compensate for the changed observation, leading to a delay or even a failure in contact event detection.
Even if a new particle set is eventually added after the delay, the existing particles may have already drifted away from the true contact point; this violates Assumption~\ref{ass:measurement_model}, thereby causing erroneous evaluations within the measurement model.
Although lowering the threshold ($\bar{\epsilon}$ in Algorithm~\ref{alg:manage_pf}) can mitigate the delay, it often induces false positives, such as initializing new particle sets in the absence of actual contact events.

\section{Mixture of Flow Matching and Particle~Filter}
\label{sec:mixture_alg}
This section presents the proposed method, \textit{ContactFlowFilter}, which integrates the \textit{flow matching} model and the \textit{particle filter}.
We employ the dual-sampling introduced in Section~\ref{subsubsec:dual_sampling_background} by incorporating the output from the flow matching model into the PF as \textit{dual particles}.

For the contact localization problem, the \textit{dual-proposal distribution} \eqref{eq:dual_proposal_def} is defined as
\begin{align}
\label{eq:dual_proposal_distribution}
    \bar{q}(\mathbf{r}_t) := \frac{p(\bm{O}_t\vert \mathbf{r}_t)}{\pi(\bm{O}_t)},
\end{align}
which can be expressed in a posterior form that enables generation via classifier-free guidance \cite{ho2022classifier} using the flow matching model \cite{frans2024one, liu2022flow}:
\begin{align}
\label{eq:cfm_posterior}
    \overline{q}(\mathbf{r}_t) 
    &= \frac{p(\bm{O}_t \vert \mathbf{r}_t)}{\int p(\bm{O}_t \vert \mathbf{r}'_t)\dd{\mathbf{r}'_t}} \nonumber
    = \frac{p(\bm{O}_t \vert \mathbf{r}_t){p(\mathbf{r}_t)}}{\int p(\bm{O}_t \vert \mathbf{r}'_t){p(\mathbf{r}'_t)}\dd{\mathbf{r}'_t}}\nonumber\\
    &= p(\mathbf{r}_t \vert \bm{O}_t).
\end{align}
Here, the prior term $p(\mathbf{r}_t)$ can be placed inside the integration assuming a non-informative prior over contact locations; i.e., contacts can occur anywhere on the robot.

The remainder of this section details the generation process from \eqref{eq:cfm_posterior} and describes how the resulting sample is weighted and incorporated into the PF.
We then introduce a robust particle set management method for contact event detection.
For brevity, when only the current time $t$ is considered, the time subscript of quantities will be omitted.

\subsection{Contact Dataset}
\label{subsec:cfm_dataset}
We first define a contact dataset $\mathcal{D}=\{ \xi_i \}^{N_d}_{i=1}$, where $N_d$ denotes the number of data points, and each data point $\xi$ consists of (i) a sample representing the contact point locations, and (ii) the corresponding observations.

\begin{figure}[t!]
\centering
    \subfloat[$n_c=1$]
    {\includegraphics[width=0.5\columnwidth, trim = {0 0px 0 0px}, clip]{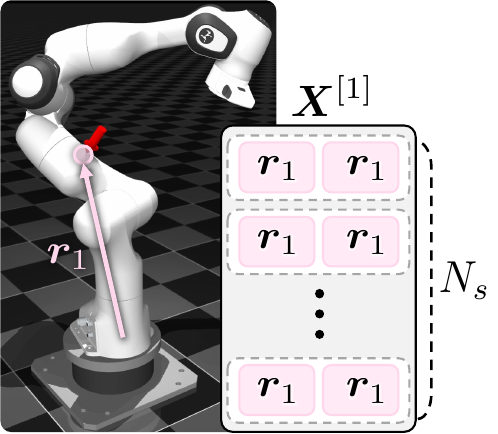}}
    \subfloat[$n_c=2$]
    {\includegraphics[width=0.5\columnwidth, trim = {0 0px 0 0px}, clip]{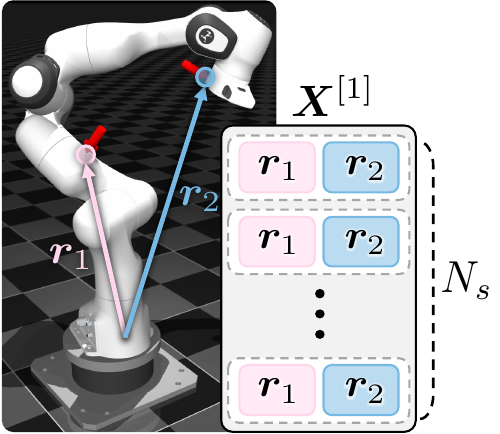}}
    \vspace{-3mm}
    \caption{
    The training sample $\bm{X}\fidx{1}\in\mathbb{R}^{N_s \times 6}$ for flow matching is constructed based on the number of contact points $n_c$:
    (a) $n_c=1$: All $N_s$ rows are filled with $\bm{r}_{1}$, extending across the full width of the matrix.
    (b) $n_c=2$: The matrix is divided into two blocks horizontally; the first three columns consist of $\bm{r}_{1}$, and the remaining three consist of $\bm{r}_{2}$.
    } 
    \vspace{-2mm}
    \label{fig:cfm_sample}
\end{figure}

\subsubsection{Sample}
\label{subsubsec:cfm_sample}
To handle our scope of up to two contacts ($n_c\in\{1,2\}$), we define a 2D data tensor $\bm{X}\fidx{1}\in\mathbb{R}^{N_{s}\times6}$, where $N_s$ denotes the number of duplicated sets of contact points.
As illustrated in Fig.~\ref{fig:cfm_sample}, when $n_c = 1$, $\bm{X}\fidx{1}$ is constructed by concatenating the first contact point $\bm{r}_{1}$ with itself to form $(\bm{r}_{1},\bm{r}_{1})\in\mathbb{R}^6$, and replicating this row $N_s$ times.
When $n_c=2$, $\bm{X}\fidx{1}$ is defined by constructing each row as $(\bm{r}_{1},\bm{r}_{2})\in\mathbb{R}^6$.

The proposed sample representation offers the following advantages:
(i) The dimension of $\bm{X}\fidx{1}$ remains constant regardless of the number of contacts $n_c$, allowing a single model to handle both single-contact and dual-contact scenarios.
(ii) It provides $N_s$ points for each contact, facilitating the representation of the contact distribution.
(iii) By learning the relationship between the pair of points within each row, the model can infer the number of contacts, enabling the generated sample to be leveraged for contact event detection (presented in Section~\ref{subsec:modified_manage_p_set}).

\subsubsection{Observation}
\label{subsubsec:cfm_obs}
\begin{figure}[t!]
\centering
    {\includegraphics[width=\columnwidth, trim = {0 0px 0 0px}, clip]{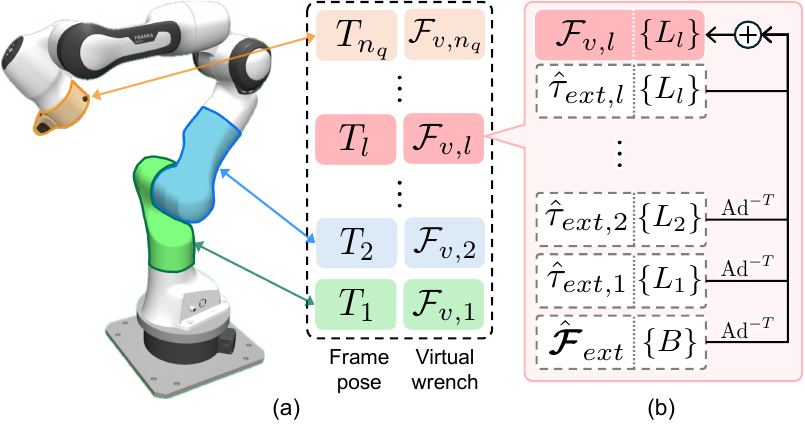}}
    \vspace{-8mm}
    \caption{ 
    (a) The observation for flow matching is defined as a sequence where each link's pose $\bm{T}_{l}$ is augmented with the virtual wrench $\mathcal{F}_{v,l}$.
    (b) $\mathcal{F}_{v,l}$ is obtained by projecting the spatially distributed external base wrench $\hat{\bm{\mathcal{F}}}_{ext}$ and the first $l$ elements of the external joint torque, $(\hat{\tau}_{ext,1}, \hat{\tau}_{ext,2}, \cdots, \hat{\tau}_{ext,l})$, onto the corresponding frame $\{L_l\}$ via the adjoint transformation $\text{Ad}^{-T}$.
    } 
    \label{fig:cfm_observation}
    \vspace{-2mm}
\end{figure}

As detailed in Section~\ref{subsec:cfm_nn_model}, the proposed method leverages a transformer-based neural network to effectively extract contact features from the observations. 
A key requirement for this architecture is a consistent input representation across both kinematic and contact-related observations. 
We express all inputs in their respective link frames (see Fig. \ref{fig:cfm_observation}), enabling the network to process them as a coherent sequence.

We represent kinematic observations using the pose of each link frame.
Specifically, $\bm{T}_{l} = \{\bm{t}_l,\bm{R}_l\}$ denotes the pose of the $l^{\text{th}}$ link frame $\fra{L_l}$, where $\bm{t}_l\in\mathbb{R}^3$ and $\bm{R}_l\in\mathbb{R}^{3\times3}$ represent the position and orientation of $\fra{L_l}$, respectively.
For contact observations, we consider the external value $\hat{\bm{\mathcal{W}}}_{ext}$.
However, its components are expressed in different reference frames; e.g., the $l^{\text{th}}$ element of the external torque $\hat{\bm{\tau}}_{ext}$ corresponds to the $z$-axis of $\fra{L_l}$, whereas the external base wrench $\hat{\bm{\mathcal{F}}}_{ext}$ is defined in the base F/T sensor frame $\{B\}$.
To achieve consistency, we align $\hat{\bm{\mathcal{W}}}_{ext}$ with each link frame by adopting the \textit{virtual wrench sensor} method, which is a modification of the approach in \cite{zwiener2018contact}.
As illustrated in Fig.~\ref{fig:cfm_observation}, the \textit{virtual wrench} at the $l^{\text{th}}$ link frame, denoted as $\bm{\mathcal{F}}_{v,l}$, is defined as
\begin{align}
\begin{split}
    \bm{\mathcal{F}}_{v,l} &= {}^{\fra{L_l}}\text{Ad}^{-T}_{\fra{B}}\hat{\bm{\mathcal{F}}}_{ext}  \\
    &+ \sum^{l}_{j=1} {}^{\fra{L_l}}\text{Ad}^{-T}_{\fra{L_j}}(0,0,0,0,0,\hat{{\tau}}_{ext,j}),
\end{split}
\end{align}
where $\hat{{\tau}}_{ext,j}$ is the $j^{\text{th}}$ element of $\hat{\bm{\tau}}_{ext}$,
and ${}^{\fra{L_l}}\text{Ad}^{-T}_{\fra{L_j}}$ is an adjoint matrix that transforms the wrench represented in the $j^{\text{th}}$ link frame $\{L_j\}$ to the $l^{\text{th}}$ link frame $\{L_l\}$.
The adjoint matrix ${}^{\fra{L_l}}\text{Ad}^{-T}_{\fra{L_j}}$ is defined as follows:
\begin{align}
   {}^{\fra{L_l}}\text{Ad}^{-T}_{\fra{L_j}} =  \begin{bmatrix}
        {}^{l}\bm{R}_{j} & 0\\
        [{}^{l}\bm{t}_{j}]_{\times}{}^{l}\bm{R}_{j} & {}^{l}\bm{R}_{j}
    \end{bmatrix},
\end{align}
where ${}^{l}\bm{R}_{j}\in\mathbb{R}^{3\times3}$ and ${}^{l}\bm{t}_{j}\in\mathbb{R}^{3}$ are the rotation matrix and translation vector from $\{L_l\}$ to $\{L_j\}$, respectively.
Intuitively, $\bm{\mathcal{F}}_{v,l}$ represents the sum of all available external wrenches acting on the predecessor links up to the $l^{\text{th}}$ link, projected onto $\{L_l\}$.
Consequently, the observation for the flow matching model is defined as $\{\bm{T}_l, \bm{\mathcal{F}}_{v,l}\}^{n_q}_{l=1}$.

\begin{rem}
\label{rem:sample_def_highlight}
    Our approach differs from \cite{cdm} primarily in the sample definition, which inherently encodes the number of contacts $n_c$.
    In \cite{cdm}, each row of a sample represents an individual contact point ($\mathbb{R}^{\bullet \times 3}$) and is processed independently by the network.
    This design makes it difficult to learn $n_c$, requiring an additional classifier to distinguish between single-contact and dual-contact scenarios.
    In contrast, our paired-point representation ($\bm{X}\fidx{1}\in\mathbb{R}^{\bullet \times 6}$) captures $n_c$, thereby eliminating the need for a separate classifier.
\end{rem}

\begin{figure*}[t]
\centering
	{\includegraphics[width=\textwidth, trim = {0 0px 0 0px}, clip]{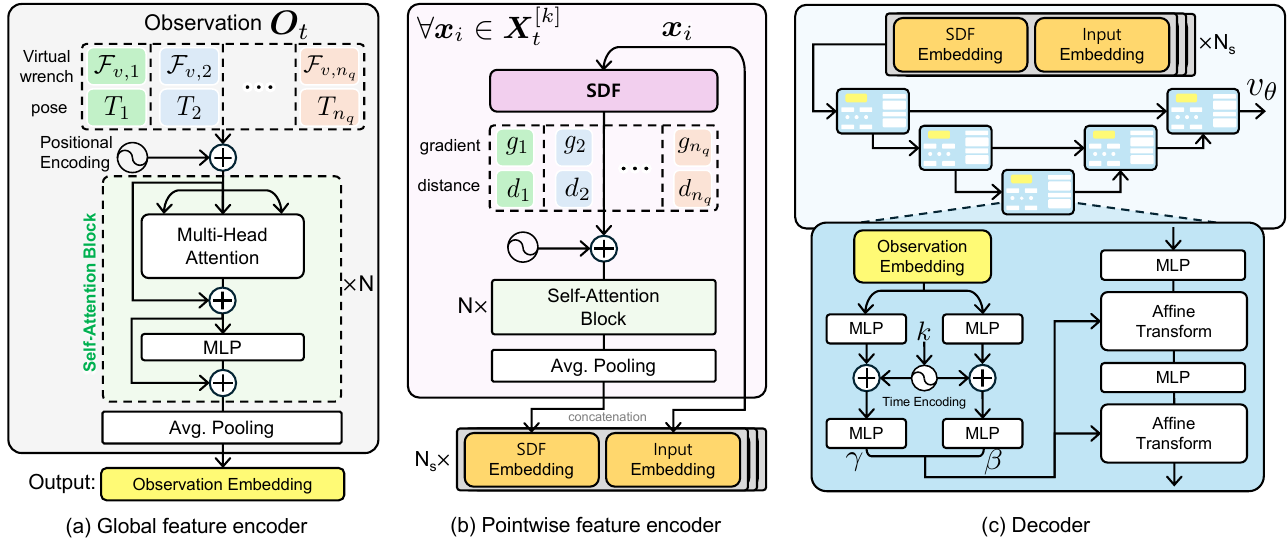}}
	\vspace{-8mm}
    \caption{
    \textbf{(a) Global feature encoder}: Encodes the correlated features of robot kinematics and external contacts.
    \textbf{(b) Pointwise feature encoder}: Captures the spatial context relative to the robot surface for each query point in $\bm{X}\fidx{k}$.
    \textbf{(c) Decoder}: Outputs the vector field by conditioning the pointwise features on the global feature.
	}
	\label{fig:cfm_architecture}
\vspace{-4mm}
\end{figure*}

\subsection{Neural Network Model}
\label{subsec:cfm_nn_model}
We rewrite the dual-proposal distribution \eqref{eq:cfm_posterior} using the definition of the sample $\bm{X}$:
\begin{align}
\label{eq:cfm_posterior_with_x}
p(\bm{X} \vert \bm{O}).
\end{align}
The first step of dual-sampling is to generate a new sample from \eqref{eq:cfm_posterior_with_x}, i.e., $\hat{\bm{X}} \sim p_{\theta}(\bm{X} \vert \bm{O})$, by integrating the corresponding ODE using the learned vector field network $\bm{v}_{\theta}(\bm{X}\vert\bm{O},k)$.
The overall structure of $\bm{v}_{\theta}$ is shown in Fig. \ref{fig:cfm_architecture}, with details provided in the remainder of this subsection.

\subsubsection{Global Feature Encoder}
\label{subsubsec:global_encoder}
As discussed in Section~\ref{subsubsec:cfm_obs}, the observations are represented with respect to each link frame ($\{\bm{T}_l, \bm{\mathcal{F}}_{v,l}\}^{n_q}_{l=1}$); i.e., the number of links $n_q$ corresponds to the \textit{sequence length}, and the feature size of each link to the \textit{embedding dimension}, as in standard Transformer input representations \cite{vaswani2017attention}.
Specifically, for the $l^{\text{th}}$ link frame, $\bm{t}_l$, $\bm{R}_l$, and $\bm{\mathcal{F}}_{v,l}$ are flattened and concatenated into a single feature vector, and these vectors are stacked across all links. 
As shown in Fig.~\ref{fig:cfm_architecture}(a), the resulting sequence is then processed through self-attention blocks to capture relationships between observations across links. 
Finally, average pooling is applied over the sequence dimension to produce a single vector, which serves as the observation embedding.

\subsubsection{Pointwise Feature Encoder}
\label{subsubsec:pointwise_encoder}
Similar to \cite{cdm}, we utilize a pre-trained SDF network \cite{rebain2021deep} to embed surface information corresponding to each point in the sample.
For notational convenience,\footnote{Although we denote the query point as $\bm{x}_i \in \mathbb{R}^3$, the actual implementation processes a 6D vector representing a pair of contact points. The pointwise feature encoder is adapted to handle this input dimension accordingly.} we will refer to $\bm{x}_i\in \mathbb{R}^{3}$ as the $i^{\text{th}}$ point in the sample $\bm{X}\fidx{k}$.
The SDF network output for the query point $\bm{x}_i$ with respect to the $l^{\text{th}}$ link surface is defined as:
\begin{align}
    (d_l, \bm{g}_l) = f_{\phi_l}(\bm{x}_i, \bm{O}) \in \mathbb{R}^{4},
\end{align}
where $d_l \!\in\! \mathbb{R}_{\geq 0}$ and $\bm{g}_l \!\in\! \mathbb{R}^{3}$ respectively denote the shortest distance and direction (unit vector) from $\bm{x}_i$ to the nearest point on the $l^{\text{th}}$ link surface, while
$\phi_l$ represents the pre-trained parameters for the $l^{\text{th}}$ link.
Consequently, the set of SDF outputs for all link surfaces is given by $\{d_l, \bm{g}_l\}_{l=1}^{n_q}$.

Similar to the global feature encoder, $\{d_l, \bm{g}_l\}_{l=1}^{n_q}$ is processed through self-attention blocks followed by average pooling (Fig.~\ref{fig:cfm_architecture}(b)).
The resulting SDF embedding is then concatenated with the input embedding of $\bm{x}_i$ to form a single vector.
Note that, as this operation is performed pointwise, it yields $N_s$ such feature vectors.

\subsubsection{Decoder}
\label{subsubsec:decoder}
Similar to U-Net \cite{ronneberger2015u}, we adopt multi-layer perceptrons (MLPs) with skip connections for the decoder structure (see Fig.~\ref{fig:cfm_architecture}(c)).
Each point embedding is conditioned with FiLM layers \cite{perez2018film}, where the observation embedding and a Fourier encoding \cite{tancik2020fourier} of the flow matching time step $k$ jointly produce the modulation parameters $\gamma$ and $\beta$.
These parameters are then applied uniformly to the pointwise feature $\bm{h}$ via the affine transformation $\gamma \odot \bm{h} + \beta$.
\begin{algorithm}
\caption{Sample-Allocation ($\hat{\bm{X}}_t,\mathfrak{X}_{t-1}$)}
\begin{algorithmic}[1]
    \preComment{initialize particle sets at the current time step}
    \State{$\mathfrak{X}_t=\emptyset$}
    \For{$\bm{x}_i$ in $\hat{\bm{X}}_t$}
        \State{$d_{\min} \gets \infty$, $j^* \gets \text{None}$}
        \For {$\mathcal{X}_{t-1,j}$ in $\mathfrak{X}_{t-1}$}
            \preComment{sample particle from previous belief}
            \State{$d \gets \|\bm{x}_i - \bm{x}^{\#}(\mathcal{X}_{t-1,j})\|$}
            \If{$d < d_{\min}$}
                \State{$d_{\min} \gets d$, $j^* \gets j$}
            \EndIf
        \EndFor 
        \preComment{add dual particle to current particle set}
        \State{\textbf{Add} $\bm{x}_i$ to $\mathcal{X}_{t,j^*}$}
    \EndFor
\State \textbf{Return} $\mathfrak{X}_{t}$
\end{algorithmic}
\label{alg:sample_allocation}
\end{algorithm}
\subsection{Dual-Sampling}
\label{subsec:mixture_dual_sampling}
As shown in \eqref{eq:cfm_posterior}, the output of the flow matching model serves as a sample from the dual-proposal distribution; i.e., $\hat{\bm{X}}_t\!\sim\! p_{\theta}(\bm{X}_t \vert \bm{O}_t) \!\approx\!\bar{q}(\mathbf{r}_t)$.
To incorporate the generated sample into the PF framework, we first assign the points in $\hat{\bm{X}}_t$ to form a new collection of particle sets $\mathfrak{X}_t$ with a size equal to that maintained by the PF.
This sample-allocation process is performed based on the distance between each point in $\hat{\bm{X}}_t$ and the particle sets from the previous time step $\mathfrak{X}_{t-1}$, as detailed in Algorithm~\ref{alg:sample_allocation}.

Recall the definition of the importance factor for dual particles given in \eqref{eq:dual_weight}.
The weight is computed by evaluating the Gaussian probability density of the motion model \eqref{eq:motion_model_gaussian_pdf}:
\begin{align}
\label{eq:dual_particle_weight}
\text{\small $p(\bm{x}_{t,i}\pidx{m}\vert\bm{x}_{t-1,i}\pidx{m})= \eta \exp \left( {-0.5\left( \norm{\bm{x}_{t,i}\pidx{m}-\bm{x}_{t-1,i}\pidx{m}}/\sigma\right)^2} \right),$}
\end{align}
where $\bm{x}_{t-1,i}\pidx{m} \sim \mathcal{X}_{t-1,i}$.
Here, $\sigma$ represents the standard deviation used in the Gaussian model of \eqref{eq:motion_model_gaussian_pdf}.
Intuitively, dual particles that are more consistent with the previous step’s particle set receive higher weights.

\subsection{Flow-Guided Particle Set Management}
\label{subsec:modified_manage_p_set}
We introduce \textit{Flow-Guided-Management} for robust contact event detection.
While Algorithm~\ref{alg:manage_pf} relies solely on the QP computation \eqref{eq:qp_after}, our approach augments this conventional criterion by integrating generative estimates $\hat{\bm{X}}_t$ from the dual-proposal distribution.

\textbf{Increasing condition:}
We combine the QP criterion with the spatial distribution of the dual particles.
While the QP condition is used with a relatively low threshold to maintain sensitivity, the dual-particle distribution is assessed subsequent to the sample-allocation process (Algorithm~\ref{alg:sample_allocation}).
Specifically, the increasing condition is satisfied only when the QP condition is met and the dual particles are widely spread, i.e., when their standard deviation within a set $\mathcal{X}_{t,i}$ exceeds that of the motion model ($\sigma$).

\textbf{Decreasing condition:} 
For particle set removal, we identify a set as a candidate if it is not assigned any particles from $\hat{\bm{X}}_t$ during the allocation process (Algorithm~\ref{alg:sample_allocation}).
If the QP-based condition supports this removal, the corresponding particle set is removed.
\begin{rem}
    To address multi-modal belief distributions, as depicted in Fig.~\ref{fig:multi_modal_qp_map}, we integrate a new condition into the conventional criteria using a logical AND.
    Conversely, combining them via a logical OR may lead to erroneous behavior.
    For instance, particles in the high-belief regions of Fig.~\ref{fig:multi_modal_qp_map}(a) might still exhibit a spread exceeding the motion model's standard deviation, causing unintended particle set initialization.
\end{rem}
\begin{algorithm}
	\caption{ContactFlowFilter ($\mathfrak{X}_{t-1}$)}
	\begin{algorithmic}[1]
	\State ${\mathfrak{X}}_t^{(1)} \leftarrow$ Sampling$(\mathfrak{X}_{t-1})$ \codeComment{in Secs.~\ref{subsec:pf_motion_model} and \ref{subsec:pf_measurement_model}}
	\State ${\mathfrak{X}}_t^{(2)} \leftarrow$ Dual-Sampling$(\mathfrak{X}_{t-1})$ \codeComment{in Sec.~\ref{subsec:mixture_dual_sampling}}
	\State $\mathfrak{X}_t \leftarrow {\mathfrak{X}}_t^{(1)} \cup {\mathfrak{X}}_t^{(2)}$ 
	\State $\mathfrak{X}_t \leftarrow$ Importance-Resampling$({\mathfrak{X}}_t)$ \codeComment{in Sec.~\ref{subsubsec:background_imp_resampling}}
	\State $\mathfrak{X}_t \leftarrow$ Flow-Guided-Management$(\mathfrak{X}_t)$ \codeComment{in Sec.~\ref{subsec:modified_manage_p_set}}
	\State \textbf{Return} $\mathfrak{X}_t$
	\end{algorithmic}
    \label{alg:contactflowfilter}
\end{algorithm}
\subsection{ContactFlowFilter}
\label{subsec:contactflowfilter}
As detailed in Algorithm~\ref{alg:contactflowfilter}, ContactFlowFilter integrates two complementary sampling strategies: standard sampling (incorporating motion and measurement models) for temporal prediction (line~1) and dual-sampling to address multi-modality (line~2).
The particle sets generated from these procedures are subsequently merged (line~3), where each unified set $\mathcal{X}_{t,i}\in\mathfrak{X}_t$ contains a varying number of standard and dual particles.
To ensure a consistent size for each particle set throughout the recursive estimation, a fixed number of particles $M$ is drawn via importance resampling (line~4). 
Finally, the number of particle sets is adjusted to match the actual number of contacts (line~5).

\section{Implementation Details}
\label{sec:imple_detail}
The proposed method, ContactFlowFilter, was implemented using both Python and C++.
The flow matching model was implemented in Python using JAX~\cite{jax2018github}, while all other components were written in C++. 
To bridge these components, we utilized Python bindings for the C++ modules.
The QP in \eqref{eq:qp_after} was solved using qpSWIFT \cite{pandala2019qpswift}.
Training and evaluation were performed on a system with an Intel Core i7-12700 CPU, an RTX 4090 GPU, and 64GB of RAM. 
The experimental setup consisted of a 7-DOF Franka Emika Panda robot arm equipped with JTSs and a base-mounted F/T sensor (ATI Omega191) (see Fig.~\ref{fig:intro_figure}).

\subsection{Dataset for Dual-Sampling}
The dataset for training the flow matching model was collected in a simulation environment using MuJoCo \cite{6386109}.
A robot with a random configuration was simulated, operating at 1000\,$\mathrm{Hz}$.
During each simulation, one or two contact forces were applied, and a data point ($\xi\in\mathcal{D}$) was collected once the system reached a stable state.
Each contact force, lying within the friction cone, had a magnitude randomly selected between 10 and 25$\,\mathrm{N}$. 
To facilitate sim-to-real transfer, noise was added to the JTSs and base F/T sensor measurements (as demonstrated in Section~\ref{subsec:exp_real}).
A total of 102.4M data points were collected, of which 80\% were used for training and the remaining 20\% for evaluation.
Empirically, we observed that constructing the dataset with 70\% dual-contact cases yielded optimal performance.

\subsection{Training Details for Dual-Sampling}
We utilized the shortcut flow matching model~\cite{frans2024one} with $K=128$ to enable fast inference with reduced ODE steps.
As demonstrated in Section~\ref{subsec:exp_selection_generative}, two ODE steps were found to be sufficient; thus, this setting was adopted in the remainder of this paper.
For stable training, the dataset was normalized to have zero mean and unit variance.
The vector field network, $\bm{v}_{\theta}$, was trained for 5M steps with a batch size of 1024, taking approximately 66 hours.
The network comprises a total of 18.1M parameters: 12.6M for the global feature encoder, 1.6M for the pointwise feature encoder, and 3.9M for the decoder (see Fig.~\ref{fig:cfm_architecture}).

\begin{figure}[t!]
\centering
    {\includegraphics[width=0.81\columnwidth, trim = {0 0px 0 0px}, clip]{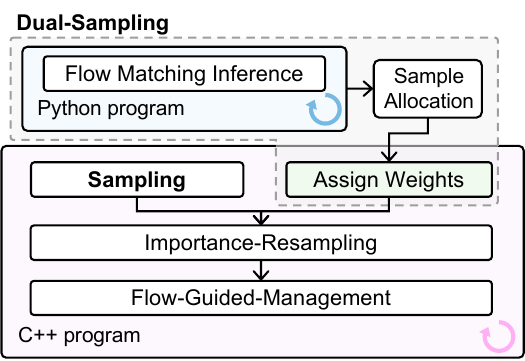}}
    \vspace{-3mm}
    \caption{
    Asynchronous execution of the flow matching inference (Python) and the main algorithm loop (C++).
    } 
    \label{fig:mixture_exec_time}
\end{figure}

\subsection{Asynchronous Sampling and Execution Time}
\label{subsec:execution_time}
In our implementation, the proposed algorithm performs sampling and dual-sampling asynchronously, as illustrated in Fig.~\ref{fig:mixture_exec_time}.
The flow matching inference is executed on a dedicated thread, separate from the main algorithm. 
The last available sample is incorporated into the PF through sample allocation and weight computation. 

We configure each particle set to contain $M=100$ particles with 64 dual particles ($N_s=32$) to balance execution time with the particle density required to represent the contact distribution.
Under these settings, the flow matching inference takes approximately 2.57\,$\mathrm{ms}$, regardless of the number of contacts.
For the main algorithmic loop, the execution time varies depending on the number of contacts: 0.69\,$\mathrm{ms}$ for the single-contact scenario and 2.69\,$\mathrm{ms}$ for the dual-contact scenario.

\section{Experimental Validation}
\label{sec:exp_val}
This section evaluates our approach in both simulated (Sections \ref{subsec:exp_selection_generative}--\ref{subsec:exp_simulation}) and real-world (Section \ref{subsec:exp_real}) environments.
For each trial across all experiments, we randomized both the robot configuration and the contact location, covering the entire surface of the manipulator.

\subsection{Validation of the Proposed Architecture for Dual-Sampling}
\label{subsec:exp_selection_generative}
In Experiment {\expgenerative}, we validate the proposed architecture (Section~\ref{subsec:cfm_nn_model}) for dual-sampling through a two-stage comparative analysis:
(i) We demonstrate its superiority by benchmarking it against a baseline method \cite{cdm}.
(ii) We evaluate Flow Matching \cite{frans2024one} against DDIM \cite{song2020denoising} to identify the superior generative framework for our architecture.
The evaluation considers both \textit{sample quality} and \textit{computational efficiency} across varying numbers of ODE integration steps.

\subsubsection{Metrics}
To quantitatively evaluate the generated samples, we employ two variants of the Root Mean Squared Error (RMSE), which we denote as Minimum-RMSE (M-RMSE) and Cluster-RMSE (C-RMSE):
\begin{align}
\label{eq:cfm_m_rmse}
    \text{\small $\textbf{M-RMSE}(\hat{\bm{X}}_t)$} &\text{\small $:= \frac{1}{n_c} \sum_{i=1}^{n_c} \min_{\forall\bm{x}_j\in\hat{\bm{X}}_t}\norm{\bm{r}_{t,i}\fidx{1} - \bm{x}_j},$} \\ 
\label{eq:cfm_c_rmse}
    \text{\small $\textbf{C-RMSE}(\hat{\bm{X}}_t)$} &\text{\small $:= \frac{1}{n_c}\sum_{i=1}^{n_c}\norm{\bm{r}_{t,i}\fidx{1}-\bm{c}_i}.$}
\end{align}
where $\bm{r}_{t,i}\fidx{1}$ is the $i^{\text{th}}$ contact point in the ground-truth sample $\bm{X}_t\fidx{1}$,
and $\bm{c}_i \in \mathbb{R}^{3}$ is the centroid of the $i^{\text{th}}$ K-means cluster of $\hat{\bm{X}}_{t}$, with the number of clusters set to $n_c$.
A low M-RMSE \eqref{eq:cfm_m_rmse} indicates that the generated samples effectively \textit{cover} the ground-truth contact points, while a low C-RMSE \eqref{eq:cfm_c_rmse} implies that the samples closely \textit{converge} to each corresponding contact point.

\begin{figure}[t]
\centering
    {\includegraphics[width=\columnwidth, trim = {0 0px 0 0px}, clip]{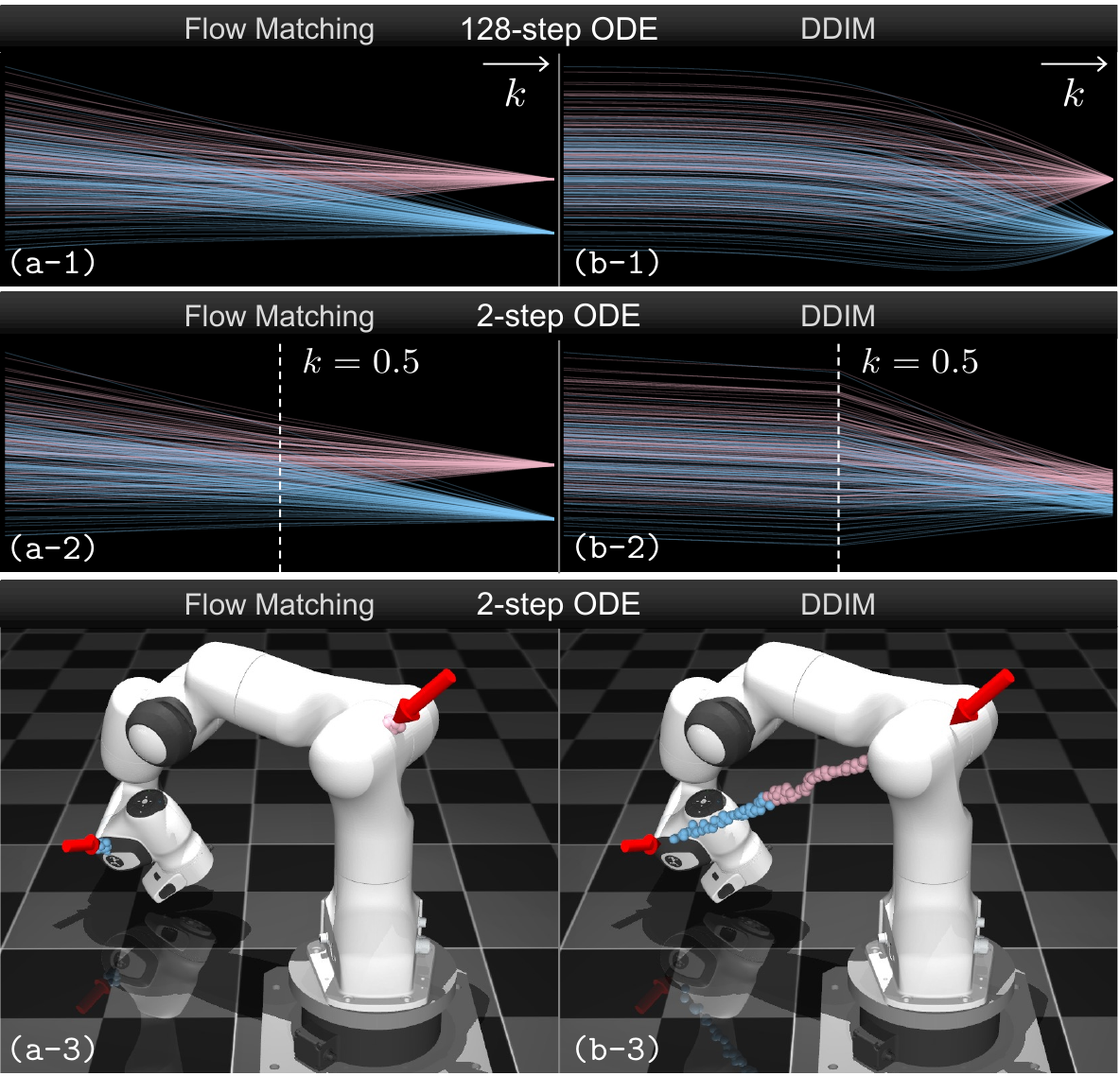}}
    \vspace{-6mm}
    \caption{
    \textbf{Experiment {\expgenerative}:}
    Comparison of generative processes implemented with our architecture, using (a) Flow Matching \cite{frans2024one} and (b) DDIM \cite{song2020denoising}.
    \textbf{\texttt{($\bullet$-1)}:} Sample trajectories visualized over ODE time steps $k$ (128-step integration).
    \textbf{\texttt{($\bullet$-2)}:} Accelerated generation paths with 2-step integration.
    \textbf{\texttt{($\bullet$-3)}:} Resulting samples via the 2-step integration.
    } 
    \label{fig:flow_proba_path}
\end{figure}

\input{table/exp_dual_sampling} 

\subsubsection{Results and Analysis}
The quantitative results are summarized in Table~\ref{tab:dual_sampling_results}.
The proposed architecture outperforms the baseline \cite{cdm} with a lower RMSE, given the same generation method (Flow Matching or DDIM).
Notably, the baseline model exhibits a high C-RMSE relative to its M-RMSE; this indicates that while the generated samples cover the contact regions, they fail to converge to the ground-truth contact points.
This limitation stems from the baseline's inability to encode the number of contacts (as discussed in Remark~\ref{rem:sample_def_highlight}), frequently resulting in misclassifications between single- and dual-contact scenarios.
Misidentifying a single-contact case as a dual one causes the generated samples to spread out rather than converge to a single location.

Within our proposed architecture, we identify Flow Matching with 2-step ODE integration as the optimal generative scheme, offering an exceptional balance between accuracy and efficiency.
As detailed in Table~\ref{tab:dual_sampling_results}, 2-step Flow Matching achieves M-RMSE values of {0.17 $\,\mathrm{cm}$} (single-contact) and {0.72 $\,\mathrm{cm}$} (dual-contact), competitively matching or even surpassing methods utilizing 16 ODE steps.
This 2-step process is eight times faster than the 16-step integration, requiring only 2.57 $\mathrm{ms}$ for inference.
Fig.~\ref{fig:flow_proba_path} visually complements these quantitative results.
The sample trajectories generated by Flow Matching (Fig.~\ref{fig:flow_proba_path}(a)) are markedly more linear than the curved paths produced by DDIM (Fig.~\ref{fig:flow_proba_path}(b)). 
This inherent linearity preserves high consistency between the results of 2-step and 128-step ODE integration.

\begin{figure}[t!]
\centering
    {\includegraphics[width=\columnwidth, trim = {0 0px 0 0px}, clip]{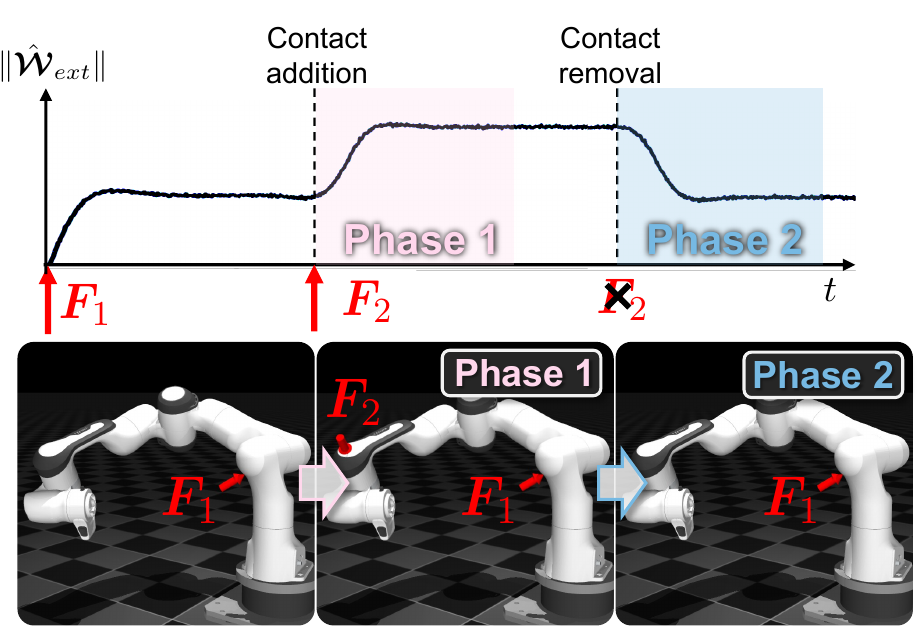}}
    \vspace{-6mm}
    \caption{
    \textbf{Experiment \expmanagepset:} 
    Evaluation of particle set management for contact event detection under sequential addition (Phase~1) and removal (Phase~2) of contact forces.
    } 
    \label{fig:exp_manage_pset_explanation}
\end{figure}

\subsection{Performance Evaluation of Particle Set Management}
\label{subsec:exp_manage_pset}
The objective of Experiment {\expmanagepset} is to compare the contact event detection performance of the proposed Flow-Guided-Management (Section~\ref{subsec:modified_manage_p_set}) against the baseline version (Algorithm~\ref{alg:manage_pf}).
To this end, we compare the proposed ContactFlowFilter (Algorithm~\ref{alg:contactflowfilter}) with the baseline CPF (Algorithm~\ref{alg:cpf_alg}).
As detailed in Section~\ref{subsec:exp_simulation}, the baseline method shows longer execution time compared to our approach.
To ensure a fair comparison that isolates the algorithmic performance, we set the execution time of the baseline to be identical to that of the proposed method for this experiment.

As illustrated in Fig.~\ref{fig:exp_manage_pset_explanation}, the experiment involves the addition and subsequent removal of a second contact force.
We define Phase~1 and~Phase 2 as the periods following the contact addition and removal, respectively.
A total of 2000 trials were conducted. 
Each trial has a duration of 3 seconds, during which a force is added at $t=1\,\mathrm{s}$ and removed at $t=2\,\mathrm{s}$.

\subsubsection{Metrics}
We evaluate the particle set management process using the following metrics:
\begin{itemize} [leftmargin=*, nosep]
    \item \textbf{Success~Rate}: During Phases 1 and 2, a trial is considered successful if the number of particle sets matches the ground-truth number of contacts.
    \item \textbf{Adjustment~Delay}: The time elapsed from the onset of a contact event (addition or removal) until the system first reaches the successful state.
    \item \textbf{Fault~Rate}: The percentage of trials exhibiting any mismatch between the number of particle sets and the ground-truth number of contacts (e.g., the erroneous initialization of a particle set in the absence of a new contact), excluding the transition period captured by the adjustment delay.
\end{itemize}

\subsubsection{Quantitative Analysis}
\input{table/exp_manage_p_sets}
The evaluation results are summarized in Table~\ref{tab:manage_pset_eval}.
The proposed method achieves a higher Success Rate, outperforming the baseline by {2.3~\%} in Phase~1 and {8.1~\%} in Phase~2.
Although the Adjustment Delay is comparable to that of the baseline in Phase~1, it is markedly reduced in Phase~2.
Most notably, the proposed method demonstrates significantly superior robustness, yielding a Fault Rate of only {0.75~\%}, in contrast to {16.5~\%} for the baseline.

\begin{figure*}[tb!]
\centering
    {\includegraphics[width=0.9 \textwidth, trim = {0 0px 0 0px}, clip]{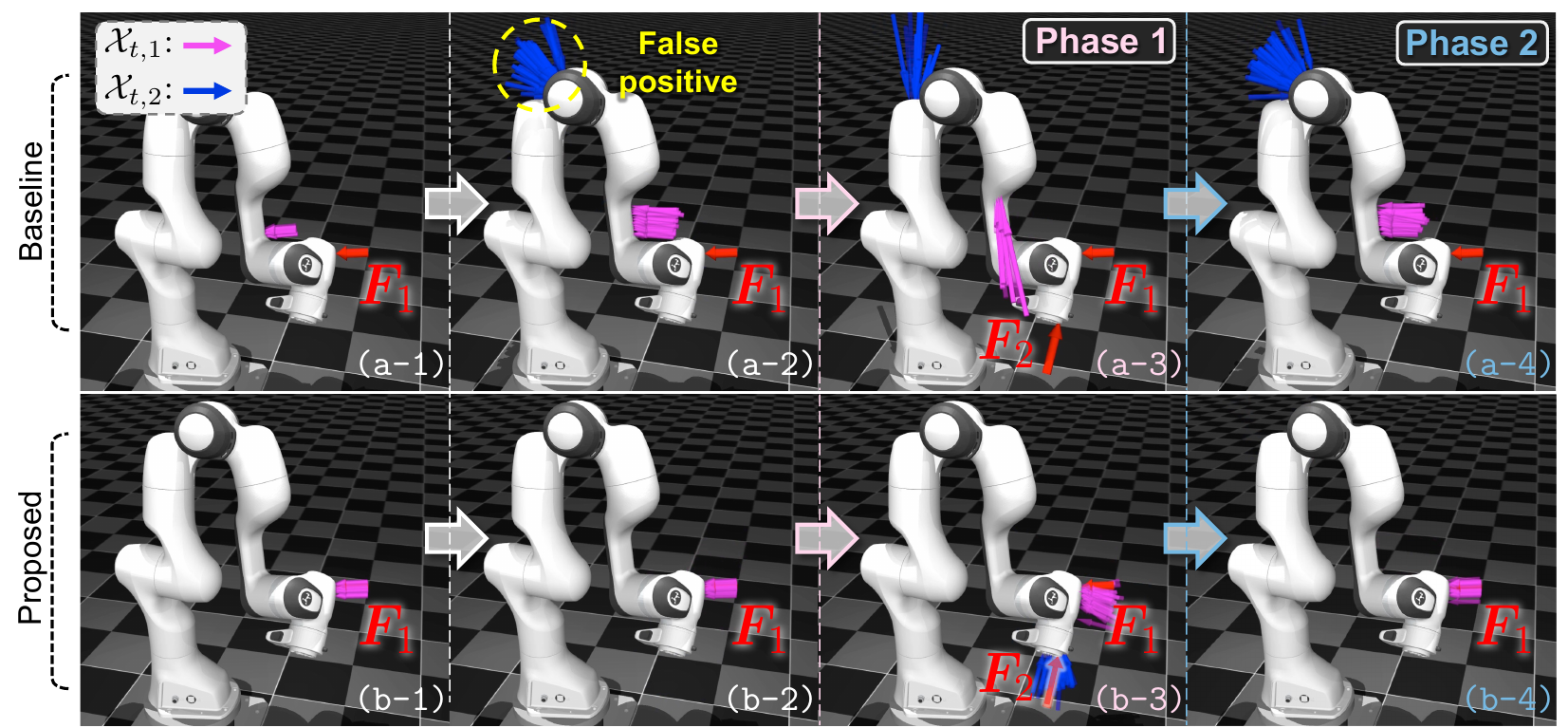}}
    \vspace{-2mm}
    \caption{
    \textbf{Experiment \expmanagepset: } 
    Sequential snapshots comparing the baseline and proposed methods. 
    (a) The baseline method suffers from erroneous set initialization due to local minima caused by particle starvation. 
    (b) The proposed method effectively performs contact event detection without such false positives.
    } 
    \label{fig:manage_p_set_first}
\vspace{-3mm}
\end{figure*}

\begin{figure*}[tb!]
\centering
    {\includegraphics[width=0.9 \textwidth, trim = {0 0px 0 0px}, clip]{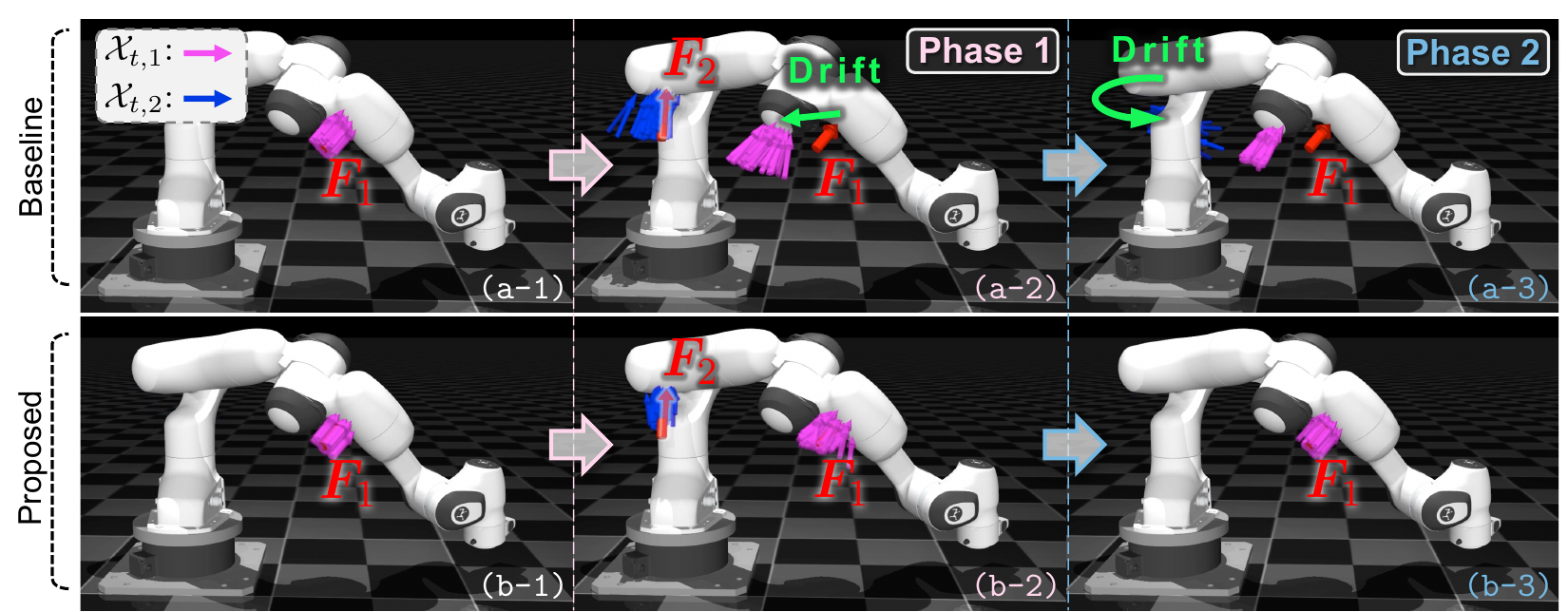}}
    \vspace{-2mm}
    \caption{
    \textbf{Experiment \expmanagepset: } 
    (a) Due to particle drift, the baseline method often fails to detect contact removal.
    (b) Conversely, the proposed method is robust against particle drift by utilizing the dual-particle distribution.
    } 
    \label{fig:manage_p_set_second}
\vspace{-3mm}
\end{figure*}

\begin{figure}[tb!]
\centering
    {\includegraphics[width=\columnwidth, trim = {0 0px 0 0px}, clip]{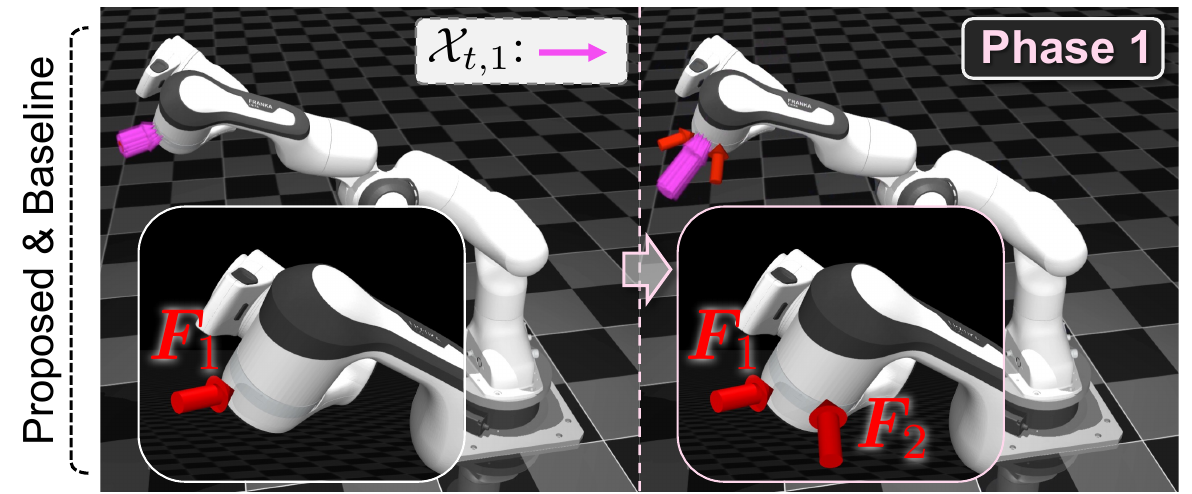}}
    \vspace{-5mm}
    \caption{
    \textbf{Experiment \expmanagepset:}
    A failure trial was observed in Phase~1 for both the baseline and proposed methods, as two adjacent contact forces are often indistinguishable from a single resultant force.
    } 
    \label{fig:manage_p_set_both_increasing_fail}
    \vspace{-2mm}
\end{figure}

\subsubsection{Qualitative Analysis}
In the following, we analyze specific failure trials observed in both the baseline and proposed methods.

\textbf{Particle starvation:}
A common failure scenario in the baseline method arises from particle starvation, where the initial particle set fails to populate high-belief regions (see Fig.~\ref{fig:manage_p_set_first}\mbox{(a-1)}).
This leads to a high QP error \eqref{eq:qp_after}, triggering the erroneous addition of a second particle set (see Fig.~\ref{fig:manage_p_set_first}\mbox{(a-2)}) and often causing a subsequent failure in Phase~2 (see Fig.~\ref{fig:manage_p_set_first}\mbox{(a-4)}). 
During the decreasing condition check (Algorithm~\ref{alg:manage_pf}), the QP error remains above the threshold even after omitting one particle set, thereby preventing the algorithm from reducing the number of particle sets.
In contrast, as shown in Fig.~\ref{fig:manage_p_set_first}(b), the proposed method avoids this issue by generating particles directly within high-belief regions.
Although this specific scenario accounts for less than {2.7~\%} of Phase~1 errors, the resulting false positive (i.e., erroneous set addition) is the primary reason of the baseline's substantially higher Fault Rate.

\textbf{Particle drift:}
The particle drift discussed in Section~\ref{subsubsec:pf_drift_v2} adversely affects the management of particle sets.
As shown in Fig.~\ref{fig:manage_p_set_second}\mbox{(a-2)}, when the number of contacts increases (Phase~1), the first particle set drifts to compensate for the new observation.
Similarly, in Phase~2 (see Fig.~\ref{fig:manage_p_set_second}\mbox{(a-3)}), drift is also observed in the second particle set.
This causes the QP-based decreasing condition (Algorithm~\ref{alg:manage_pf}) to fail, as the algorithm cannot determine which particle set to remove.
In contrast, the proposed method directly generates dual particles, regardless of any particle drift (see Fig.~\ref{fig:manage_p_set_second}(b)).
By leveraging the distribution of these dual particles for set management, our approach demonstrates robustness against particle drift.

\textbf{Ambiguity of Adjacent Contacts:}
Due to the sparsity of proprioceptive sensors, two contact forces with similar locations and directions are often indistinguishable from a single resultant force acting at their center (see Fig.~\ref{fig:manage_p_set_both_increasing_fail}).
This inevitable ambiguity leads to estimation failures for both the baseline and the proposed methods, as particles converge to a single point.
In the proposed method, this ambiguity is the primary cause of failure, with approximately {90.9~\%} of the failures in Phase~1 occurring when the two contacts are located on adjacent links.

\subsection{Comparative Study of Contact Estimation Algorithms}
\label{subsec:exp_simulation}
In Experiment {\expbelief}, we conduct a comparative evaluation to assess the contact estimation performance of the proposed \textit{ContactFlowFilter}.
Our evaluation focuses on two primary aspects: the \textit{dissimilarity} between the particle distribution and the ground-truth distribution, and the \textit{estimation error} with respect to the ground-truth contact state ($\bm{r}_{t,i}$ and $\bm{F}_{t,i}$).
As in the previous section, we consider a sequential contact scenario, i.e., the period until the end of Phase~1 (see Fig.~\ref{fig:exp_manage_pset_explanation}).
We benchmark our approach against the following three baseline methods:
\begin{itemize}[leftmargin=*, nosep]
\item \textbf{CPF} \cite{manuelli2016localizing}:
This baseline is identical to Algorithm~\ref{alg:cpf_alg} except for its motion model.
Instead of using pre-processed mesh data, the motion model first adds 3D Gaussian noise to the particles and then projects them onto the robot's surfaces.
This projection step introduces computational overhead, increasing the execution time to 33.93 $\mathrm{ms}$ and 69.08 $\mathrm{ms}$ for single- and dual-contact scenarios, respectively.
\item \textbf{MCP-EP} \cite{10161173}:
In this baseline, the dual-sampling step in Algorithm~\ref{alg:contactflowfilter} is replaced with a random sampling strategy.
When the current particle set fails to explain the observation (as indicated by a high QP error), the algorithm generates random particles to explore neighboring links.
Its execution time is identical to that of the proposed method: 0.69$\,\mathrm{ms}$ and 2.69$\,\mathrm{ms}$ for single-contact and dual-contact scenarios, respectively.
\item \textbf{Dual-Particle only}:
This baseline utilizes samples generated via the flow matching model.
To enable comparison with PF-based approaches, we use the dual particles for this baseline, i.e., the particles allocated to the particle set (output of Algorithm~\ref{alg:sample_allocation}).
\end{itemize}

\subsubsection{Metrics}
We present the metrics used to evaluate dissimilarity and estimation error.

\begin{figure}[t!]
\centering
    {\includegraphics[width=0.9\columnwidth, trim = {0 0px 0 0px}, clip]{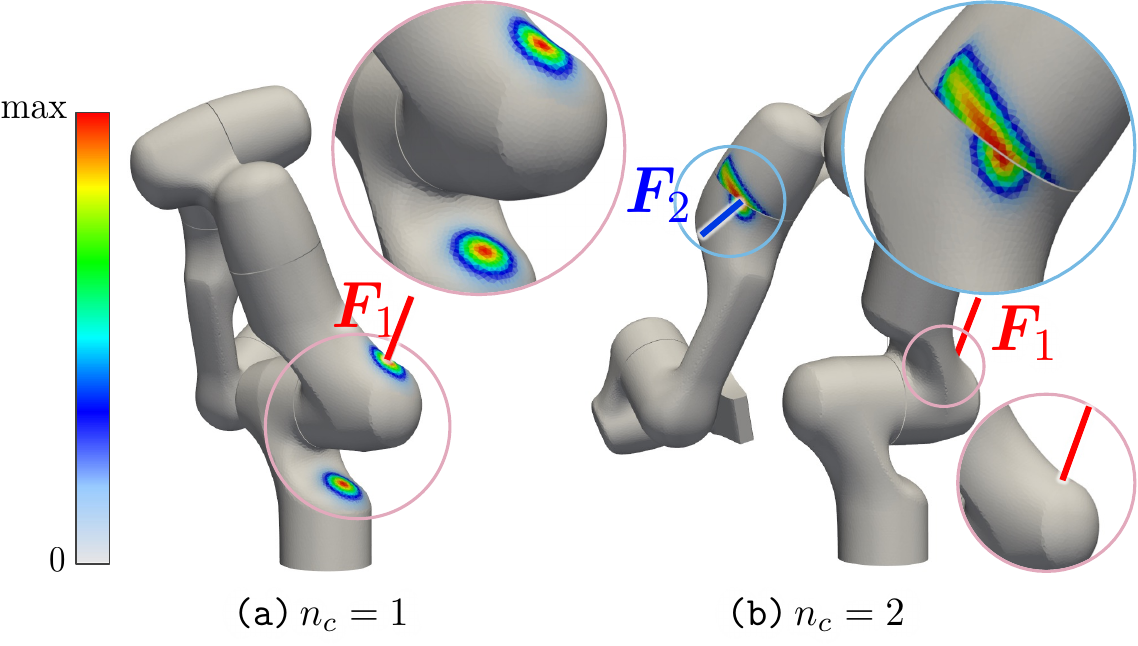}}
    \vspace{-3.5mm}
    \caption{
    \textbf{Experiment {\expbelief}:}
    The likelihood function, defined in \eqref{eq:likelihood_def}, is visualized over the entire robot surface for two scenarios: (a) single-contact ($n_c=1$) and (b) dual-contact ($n_c=2$).
    In (b), the likelihood is computed for the second contact, resulting in a negligible value at the first contact point (indicated by pink circle).
    }
    \label{fig:likelihood_vis}
\end{figure}

\textbf{Dissimilarity:}
Computing the dissimilarity requires the ground truth of the target distribution \eqref{eq:target_posterior}.
We construct this distribution by defining a likelihood function derived from the QP \eqref{eq:qp_after}.
For sequential contact scenarios, the likelihood function is defined as follows (see Fig.~\ref{fig:likelihood_vis} for a visualization):
\begin{align}
\label{eq:likelihood_def}
\small
    \!\!L( \bm{O}_{0:t} \vert & \bullet)
    \!:=\!\!
    \begin{cases}
        \eta \exp \left(-\alpha QP(\bm{O_t}\vert \{\bullet\}) \right)
         &\!\!\!\!\!{\scriptsize \text{if } \text{$n_c = 1$}}
         \\
         \eta \exp \left(-\alpha QP(\bm{O_t}\vert \{ \bm{r}_{t,1}, \bullet\}) \right)
         &\!\!\!\!\!{\scriptsize \text{if } \text{$n_c = 2$}}
    \end{cases}
    .
\end{align}
For a single contact ($n_c=1$), the QP is computed for each face of the pre-processed mesh.
For dual contacts ($n_c=2$), the location of the first contact is fixed at its ground-truth value, while the second contact is computed over the entire robot surface.
Notably, since the dual-contact scenario incorporates the first contact as known historical information in \eqref{eq:likelihood_def}, this setup allows us to assess the algorithm's capability to handle temporal dependencies related to discrete state evolution.

\input{table/exp_belief}

Based on \eqref{eq:likelihood_def}, we employ two metrics to quantify the dissimilarity between distributions: the Particle log-Likelihood Score (PLS) and the Earth Mover's Distance (EMD).
The PLS is defined as the average negative log-likelihood of the particles:
\begin{align}
\label{eq:metric_pls}
\textbf{PLS}:=\frac{1}{M}\sum^{M}_{m=1}-\log \left( L(\bm{O}_{0:t} \vert \bm{x}_{t,n_c}\pidx{m} ) \right).
\end{align}
Here, the PLS is evaluated for particles $\bm{x}_{t,n_c}\pidx{m} \!\in\! \mathcal{X}_{t,n_c}$ corresponding to the final index $n_c$.
The EMD measures the minimum cost required to transform one distribution into another by solving the following optimization problem:
\begin{subequations}
\label{eq:emd} 
\begin{align}
    \textbf{EMD}&:=\min_{T\in\mathbb{R}^{M\times N_f}}  \sum_{n=1}^{N_f} \sum_{m=1}^{M} T_{n,m} \cdot \norm{\bm{y}\pidx{n}-\bm{x}_{t,n_c}\pidx{m}} \label{eq:emd_objective} \\
    \text{s.t.} \quad  &\sum_{n=1}^{N_f} T_{n,m} = \frac{1}{M}, \quad \quad \quad \quad \quad \, {\scriptstyle \forall m \in \{1,2, \cdots,M \}}, \label{eq:ocp_cons1} \\
    & \sum_{m=1}^{M} T_{n,m} = L ( \bm{O}_{0:t} \vert \bm{y}\pidx{n}), \quad {\scriptstyle \forall n \in \{1,2, \cdots,N_f \}}, \label{eq:ocp_cons2}
\end{align}
\end{subequations}
where $\bm{y}\pidx{n}$ represents the centroid of the $n^{\text{th}}$ face of the $N_f$ faces in the mesh.
Here, we assume that all particles have uniform weights, as this stage follows the resampling process.

While PLS is effective for verifying that particles are concentrated in high-likelihood regions, it is limited in multi-modal cases.
It indicates good performance even when particles converge to only a single mode (e.g., the multi-modal case in Fig.~\ref{fig:likelihood_vis}(a)).
In contrast, EMD measures the transport cost; thus, particles must populate all modes to minimize this metric.

\begin{figure}[t!]
\centering
    {\includegraphics[width=\columnwidth, trim = {0 0px 0 0px}, clip]{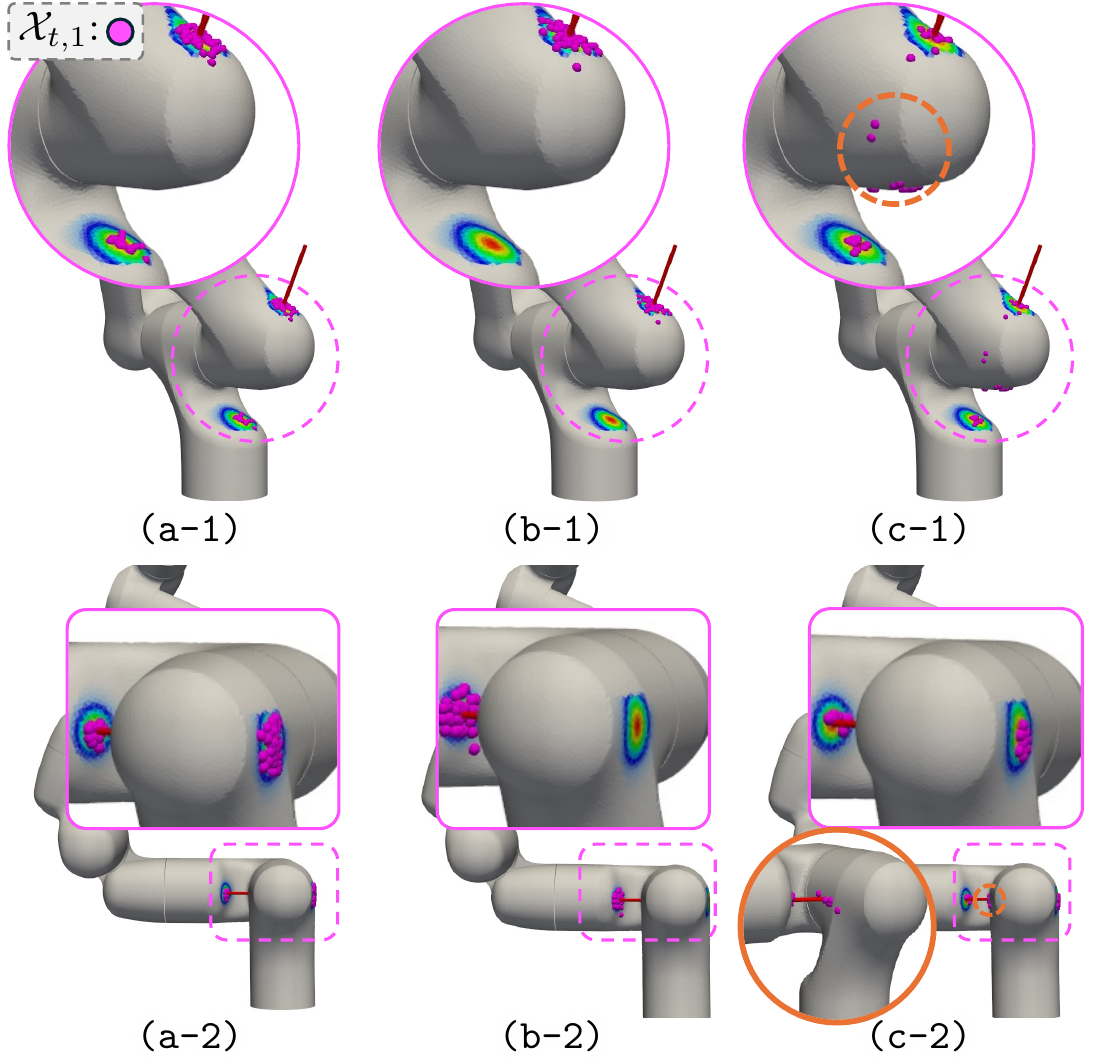}}
    \vspace{-7mm}
    \caption{
    \textbf{Experiment {\expbelief}:}
    Comparison of three methods in multi-modal single-contact scenarios: (a) ContactFlowFilter, (b) CPF, and (c) Dual-Particle-Only.
    }
    \label{fig:exp_c_multi_modal}
\end{figure}

\begin{figure}[t!]
\centering
    {\includegraphics[width=\columnwidth, trim = {0 0px 0 0px}, clip]{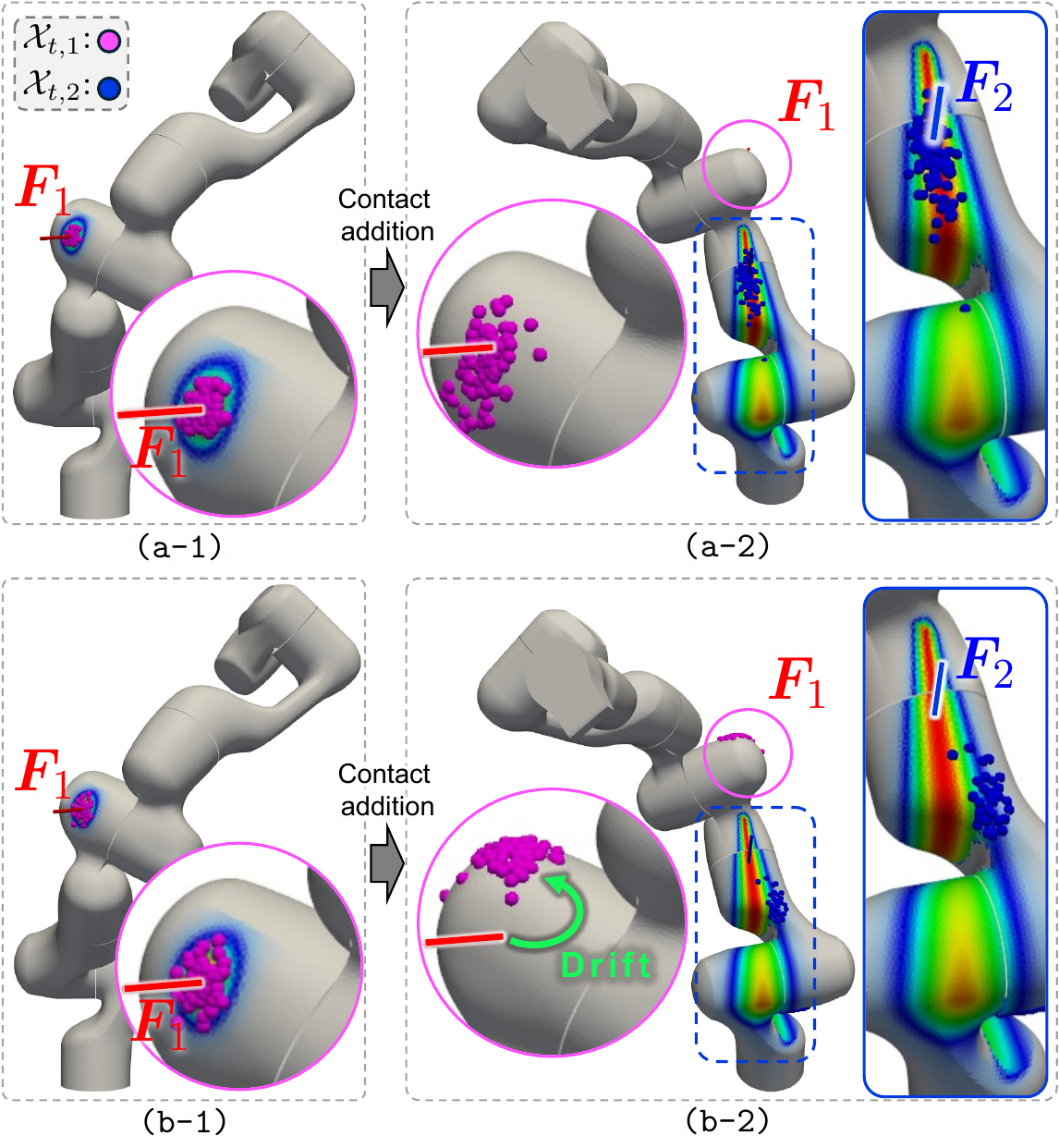}}
    \vspace{-8mm}
    \caption{
    \textbf{Experiment {\expbelief}:}
    The first row (a-$\bullet$) presents the results of ContactFlowFilter, and the second row (b-$\bullet$) shows those of CPF.
    Each row illustrates the sequential contact scenario, evolving from a single-contact ($n_c=1$) to a dual-contact ($n_c=2$).
    }
    \label{fig:exp_c_drift}
    \vspace{-4mm}
\end{figure}

\textbf{Estimation error:}
As in Section~\ref{subsec:exp_selection_generative}, we report the Minimum-RMSE (M-RMSE) and Weighted-RMSE (W-RMSE) for the particle sets $\mathfrak{X}_t$:
\begin{align}
\label{eq:metric_m_rmse}
\text{\small $\textbf{M-RMSE}(\mathfrak{X}_t)$} &\text{\small $:= \frac{1}{n_c}\sum_{\mathcal{X}_{t,i}\in\mathfrak{X}_t}\min_{\forall\bm{x}_{t,i}\pidx{m}\in\mathcal{X}_{t,i}}\norm{\bm{r}_{t,i} - \bm{x}_{t,i}\pidx{m}},$}\\
\text{\small $\textbf{W-RMSE}(\mathfrak{X}_t)$}&\text{\small $:= \frac{1}{n_c}\sum_{\mathcal{X}_{t,i}\in\mathfrak{X}_t}\norm{\bm{r}_{t,i} - \bm{x}^{*}(\mathcal{X}_{t,i})}$}.
\end{align}
The W-RMSE quantifies the error of the particle with the highest weight ($\bm{x}^{*}(\mathcal{X}_{t,i})$) relative to the ground truth.
We also apply the same M-RMSE and W-RMSE metrics to contact forces.

\subsubsection{Quantitative Analysis}
The results presented in Table~\ref{tab:exp_belief} indicate that the proposed method outperforms the baselines across all metrics.
Our approach achieves the lowest dissimilarity from the ground-truth target distribution, as evidenced by the lowest PLS and EMD values.
It also achieves the lowest RMSE for both contact point and force estimation.
\input{table/exp_real}

\subsubsection{Qualitative Analysis}
Fig.~\ref{fig:exp_c_multi_modal} illustrates scenarios involving multi-modal contact distributions in a single-contact case.
As depicted in the first column (Fig.~\ref{fig:exp_c_multi_modal}\mbox{(a-$\bullet$)}), the proposed method successfully distributes particles across both high-belief regions.
In contrast, the CPF method, shown in the second column (Fig.~\ref{fig:exp_c_multi_modal}\mbox{(b-$\bullet$)}), fails to capture this multi-modality, with its particles converging to a single region.
This limitation results in a significantly higher EMD value, indicating a substantial dissimilarity from the ground-truth distribution.
Moreover, as shown in the third column (Fig.~\ref{fig:exp_c_multi_modal}\mbox{(c-$\bullet$)}), certain particles generated by the flow matching model are located outside the high-belief contact regions (highlighted in orange).
This occurs because the flow matching model predicts contact points in 3D space rather than being strictly constrained to the robot surface.
Even when incorporated into the PF framework, these outliers are effectively suppressed by assigning low weights during (i) the dual-sampling step (due to inconsistency) and (ii) the QP-based measurement update.

Fig.~\ref{fig:exp_c_drift} demonstrates a failure case caused by particle drift (discussed in Section~\ref{subsubsec:pf_drift_v2}) in sequential contact estimation.
The results of the proposed method and CPF are presented in Fig.~\ref{fig:exp_c_drift}\mbox{(a-$\bullet$)} and Fig.~\ref{fig:exp_c_drift}\mbox{(b-$\bullet$)}, respectively.
While both methods initially achieve accurate estimation for the first contact, the first particle set in the baseline method begins to drift when the second contact occurs (see Fig.~\ref{fig:exp_c_drift}\mbox{(b-2)}).
Since CPF relies on the measurement model that updates weights assuming other particle sets have converged to their true contact points (Assumption~\ref{ass:measurement_model}), this drift leads to a cascading error.
As a result, the second particle set fails to populate the high-belief region of its target distribution.

\begin{figure}[t!]
\centering
    {\includegraphics[width=\columnwidth, trim = {0 0px 0 0px}, clip]{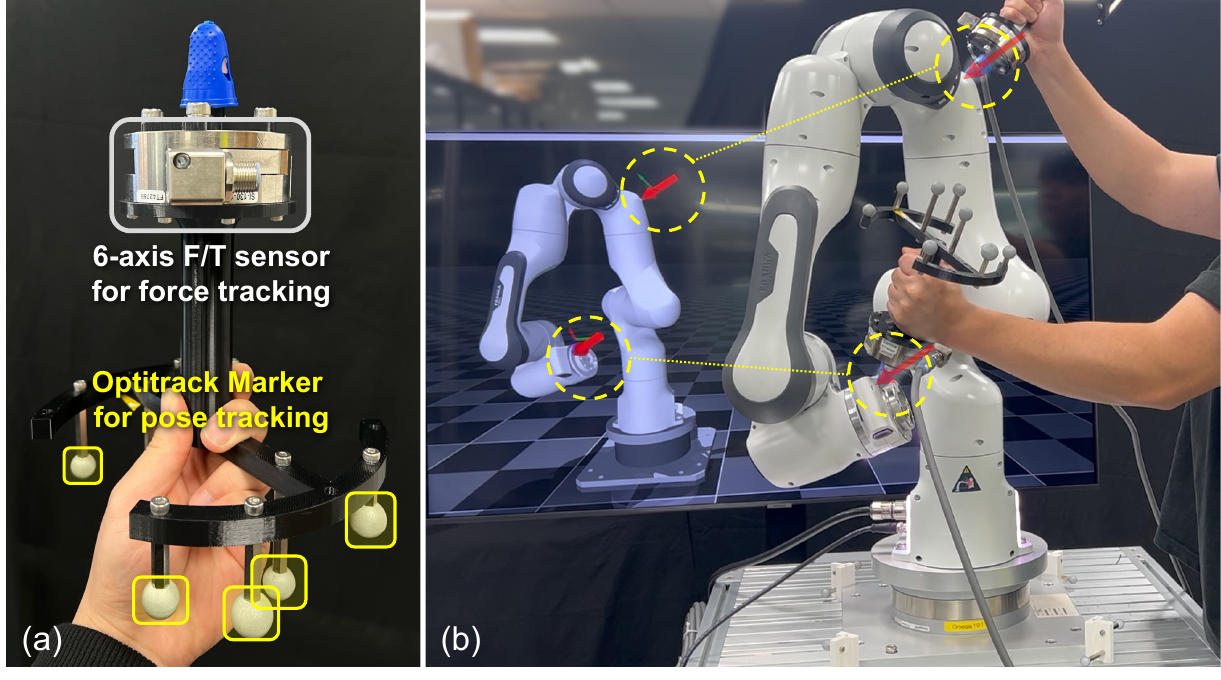}}
    \vspace{-6mm}
    \caption{
    \textbf{Experiment {\expreal}:}
    The real-world evaluation dataset was collected (a) using a tool for acquiring ground-truth contact states ($\bm{r}_i$ and $\bm{F}_i$), (b) under a sequential dual-contact scenario.
    }
    \label{fig:exp_d_data_collection}
\end{figure}

\subsection{Real-World Experiment}
\label{subsec:exp_real}
Experiment {\expreal} validates the proposed method in a real-world setting.
We employed the same evaluation metrics and experimental scenarios as in the simulation experiments detailed in Section~\ref{subsec:exp_simulation}.
For dual-sampling, the flow matching model trained in simulation was directly transferred to the physical robot.

To assess the contact estimation algorithms under consistent real-world contact scenarios, we collected a dedicated evaluation dataset (i.e., not for training).
We used a specialized tool, shown in Fig.~\ref{fig:exp_d_data_collection}(a), to acquire ground-truth contact locations ($\bm{r}_i$) and forces ($\bm{F}_i$).
As illustrated in Fig.~\ref{fig:exp_d_data_collection}(b), this tool was used to apply two sequential contacts to the robot while recording the corresponding observations.
A total of 200 trials were conducted across 20 distinct robot configurations.

The quantitative results of the real-world experiment are summarized in Table~\ref{tab:exp_real}.
Consistent with the simulation results, the proposed method demonstrates superior performance, even in the real-world setting.
The supplementary video presents the experimental procedure and qualitative analysis, along with additional contact scenarios (e.g., sliding contact shown in Fig.~\ref{fig:challenges}(b) and contact during robot motion).

\section{Summary, Conclusion, and Discussion}
\label{sec:conclusion}
In this paper, we propose a novel method called \textit{ContactFlowFilter} for proprioception-based online contact estimation.
Our method integrates a generative model with a particle filter (PF) by adopting a dual-sampling concept.
In this framework, particles are drawn from a mixture of two proposal distributions: (i) ``standard particles'' from the PF's prediction step and (ii) ``dual particles'' from the generative model.
The mixture proposal enables ContactFlowFilter to simultaneously tackle key challenges in contact estimation: \textit{multi-modality}, \textit{temporal dependency}, and \textit{contact event detection}.

To address \textit{temporal dependency}, we utilize the PF framework as the foundation of our approach, leveraging its recursive nature to handle time-dependent belief distributions.
However, PFs alone encounter two fundamental limitations that restrict their ability to address the remaining challenges.
To handle \textit{multi-modality}, we resolve ``particle starvation'' (Section~\ref{subsubsec:pf_starvation}) by incorporating dual particles from the generative model.
This dual-sampling step enables the direct generation of samples in high-belief regions, even where the particles fail to populate.
To mitigate the sensitivity of conventional criteria for \textit{contact event detection}, we propose a method leveraging generative estimates. 
Our approach maintains robustness against ``particle drift'' (Section~\ref{subsubsec:pf_drift_v2}), which typically leads to contact state tracking failure; e.g., significant errors in localization or erroneous contact event detection.

To implement our dual-sampling strategy, we employ a flow matching model, specifically exploiting its linearized probability path.
This feature enables rapid state estimation, critical for high-fidelity temporal tracking, while remaining highly effective at capturing the multi-modal distribution of contact states.
Such capabilities are underpinned by our novel architecture, which incorporates a spatially aligned data representation and a tailored transformer-based neural network.

We validate \textit{ContactFlowFilter} through extensive comparative studies in both simulation and real-world settings. 
Quantitative and qualitative analyses confirm that the proposed method effectively resolves the aforementioned challenges.
We validate the proposed generative architecture for dual-sampling (Section~\ref{subsec:exp_selection_generative}) and the robustness of our method in handling discrete contact transitions (Section~\ref{subsec:exp_manage_pset}).
Ultimately, our approach outperforms the baselines in contact estimation, achieving the highest accuracy and the lowest dissimilarity to the target posterior (Sections~\ref{subsec:exp_simulation} and \ref{subsec:exp_real}, and supplementary video).

Despite the promising results, the current implementation of our method is limited to estimating a maximum of two contacts.
This constraint stems from the generative component, which relies on a fixed-size array of training samples with up to two contacts (Section~\ref{subsubsec:cfm_sample}).
Although it readily scales to additional contacts, we constrained the scope to isolate the core challenges, particularly \textit{multi-modality}. 
For up to two contacts, the generative model captures multi-modality within the solution space bounded by the base F/T sensor.
Beyond two, estimating multiple contacts from single observations becomes ill-posed, as the solution space for the time-independent model contains numerous distinct contact states yielding identical sensor measurements.
We plan to extend the method to multi-contact scenarios by incorporating additional sensing modalities (e.g., F/T sensors or a dual-arm setup) or by conditioning the generative component on temporal context to disambiguate the solution space.

%% file: table/related_works_method_comparison.tex
\begin{table*}[t]
\caption{Comparison of different approaches based on key capabilities.}
\centering
{
\renewcommand{\tabularxcolumn}[1]{>{\centering\arraybackslash}m{#1}}
\scriptsize

\begin{tabularx}{\textwidth}{ l *{7}{X} >{\hsize=0.8\hsize}X }
\toprule
 & \multicolumn{3}{c}{\textbf{Model-based Approaches}} & \multicolumn{4}{c}{\textbf{Learning-based Approaches}} & \multirow{2}{*}{\textbf{Ours}} \\ 
\cmidrule(lr){2-4} \cmidrule(lr){5-8} 
 & { Line of force} & { Particle filter} & { QP-based algorithm} & { Probabilistic mapping} & { Binary classification} & { Regression} & { Diffusion model} & \\ 
\midrule

\mbox{Contact Number Support} & $1$
 & $\geq2$ & $1$  & $\geq2$ & $1$ & $1$  & $2$ & $2$ \\

\mbox{Multi-Modal Distribution} & $\times$
  & $\times$  & $\bigcirc$ & $\bigcirc$ & $\times$ & $\times$  & $\bigcirc$ & $\bigcirc$ \\

\mbox{Temporal Reasoning}       & $\times$
  & $\bigcirc$ & $\times$  & $\times$  & $\times$ & $\times$  & $\triangle$ & $\bigcirc$ \\

\mbox{Contact Force Estimation} & $\bigcirc$
 & $\bigcirc$ & $\bigcirc$ & $\times$  & $\times$ & $\times$  & $\times$ & $\bigcirc$ \\
\cmidrule(lr){1-9}

\mbox{References}
& \cite{buondonno2016combining, 8059840, iskandar2021collision, popov2022analytical}
& \cite{10342345, manuelli2016localizing, 10161173, zwiener2019armcl, zurlo2023collision, bimbo2019collision, popov2021multi, job2023multiple} 
& \cite{pang2021identifying, popov2021real} 
& \cite{liang2021contact} 
& \cite{zwiener2018contact, popov2020transfer} 
& \cite{popov2020transfer, fu2025unitac} 
& \cite{cdm} 
& -- \\ 
\bottomrule
\end{tabularx}
} 
\vspace{-4mm}
\label{tab:method_review}
\end{table*}

%% file: table/exp_dual_sampling.tex
\begin{table*}[t]
\caption{\textbf{Experiment \expgenerative:} Quantitative results of samples generated via dual-sampling (inference time: approx. 1.28 $\mathrm{ms}$/step)}
\scriptsize
\centering
\begin{tabularx}{\textwidth}{*{7}{>{\centering\arraybackslash}X}}
\toprule
\multirow{2}{*}{Generation method} &
\multirow{2}{*}{\shortstack{Architecture}} &
  \multirow{2}{*}{ODE-step} &
  \multicolumn{2}{c}{Single-contact ($n_c=1$)} &
  \multicolumn{2}{c}{Dual-contact ($n_c=2$)} \\
  \cmidrule(l){4-7}
 & & & M-RMSE ($\mathrm{cm}$) $\downarrow$ & C-RMSE ($\mathrm{cm}$) $\downarrow$ & M-RMSE ($\mathrm{cm}$) $\downarrow$ & C-RMSE ($\mathrm{cm}$) $\downarrow$ \\
\midrule
\multirow{4}{*}{\shortstack{Flow Matching}} & 
\multirow{4}{*}{\ding{51} Proposed}
   & 2   & 0.17 ± 0.27 & 0.45 ± 0.63 & 0.72 ± 0.92 & 2.20 ± 2.75  \\
 &  & 4   & 0.16 ± 0.27 & 0.44 ± 0.62 & 0.67 ± 0.89 & 2.18 ± 2.86  \\
 &  & 8   & 0.15 ± 0.26 & 0.43 ± 0.62 & 0.62 ± 0.86 & 2.17 ± 2.91  \\
 &  & 16  & 0.14 ± 0.26 & 0.42 ± 0.62 & 0.56 ± 0.85 & 2.16 ± 2.94  \\
\midrule
\multirow{4}{*}{\shortstack{Flow Matching}} &
\multirow{4}{*}{\ding{55} Baseline}
   & 2   & 0.40 ± 0.51 & 1.36 ± 1.88 & 1.17 ± 1.15 & 3.88 ± 3.59  \\
 &  & 4   & 0.31 ± 0.44 & 1.20 ± 1.85 & 0.92 ± 1.05 & 3.60 ± 3.79  \\
 &  & 8   & 0.25 ± 0.42 & 1.09 ± 1.80 & 0.77 ± 1.00 & 3.46 ± 3.85  \\
 &  & 16  & 0.21 ± 0.41 & 1.01 ± 1.74 & 0.69 ± 0.98 & 3.36 ± 3.87  \\
\midrule
\multirow{4}{*}{\shortstack{DDIM}} &
\multirow{4}{*}{\ding{51} Proposed}
   & 2   & 0.26 ± 0.53 & 0.60 ± 0.72 & 6.13 ± 2.50 & 13.63 ± 5.12 \\
 &  & 4   & 0.21 ± 0.36 & 0.53 ± 0.71 & 1.54 ± 1.62 & 6.03 ± 2.39  \\
 &  & 8   & 0.19 ± 0.29 & 0.50 ± 0.71 & 1.05 ± 1.22 & 2.86 ± 2.87  \\
 &  & 16  & 0.18 ± 0.26 & 0.50 ± 0.72 & 0.85 ± 1.04 & 2.43 ± 3.07  \\
\midrule
\multirow{4}{*}{\shortstack{DDIM}} &
\multirow{4}{*}{\ding{55} Baseline}
   & 2   & 0.73 ± 1.02 & 1.46 ± 1.45 & 8.80 ± 2.92 & 15.09 ± 6.11 \\
 &  & 4   & 0.43 ± 0.83 & 1.04 ± 1.38 & 2.32 ± 1.85 & 7.25 ± 3.08  \\
 &  & 8   & 0.32 ± 0.60 & 0.89 ± 1.35 & 1.33 ± 1.32 & 3.69 ± 3.40  \\
 &  & 16  & 0.26 ± 0.46 & 0.82 ± 1.36 & 0.96 ± 1.08 & 3.15 ± 3.61  \\
\bottomrule
\end{tabularx}
\vspace{-4mm}
\label{tab:dual_sampling_results}
\end{table*}

%% file: table/exp_manage_p_sets.tex
\begin{table}[t]
\caption{\textbf{Experiment {\expmanagepset}: } Quantitative results of particle set management}
\centering
\scriptsize
\begin{tabularx}{\linewidth}{c *{5}{>{\centering\arraybackslash}X}}
\toprule
& \multicolumn{2}{c}{Suc. Rate $\uparrow$} &
  \multicolumn{2}{c}{Adj. Delay $\downarrow$} &
  \multicolumn{1}{c}{Fau. Rate $\downarrow$} \\
\cmidrule(lr){2-3}\cmidrule(lr){4-5}\cmidrule(lr){6-6}
\multicolumn{1}{c}{Method} &
  \shortstack{Phase 1\\(\%)} & \shortstack{Phase 2\\(\%)} &
  \shortstack{Phase 1\\(ms)} & \shortstack{Phase 2\\(ms)} &
  \shortstack{Overall\\(\%)} \\
\midrule
\ding{51}\ Proposed & \textbf{99.60} & \textbf{99.65} & {3.21} & \textbf{26.74} & \textbf{0.75} \\
\addlinespace 
\ding{55}\ Baseline & 97.30 & 91.55 & \textbf{2.89} & 51.46 & 16.50 \\
\bottomrule
\end{tabularx}
\label{tab:manage_pset_eval}
\vspace{-4mm}
\end{table}

%% file: table/exp_belief.tex
\begin{table*}[!ht]
\caption{\textbf{Experiment {\expbelief}:} Quantitative results of the simulation evaluation}
\scriptsize
\centering
    \sisetup{
      table-align-uncertainty = true,
      separate-uncertainty = true,
      detect-weight = true,  
      mode = text,
    }
    \renewcommand{\arraystretch}{1.1} 
    
    \begin{tabular*}{\textwidth}{
      @{\extracolsep{\fill}} 
      l 
      l 
      S[table-format=2.3(3.2)] 
      S[table-format=2.3(3.2)] 
      S[table-format=2.3(3.2)] 
      S[table-format=2.3(3.2)] 
      @{}
    }
    \toprule
    \multirow{2}{*}{\parbox{2.0cm}{\centering {Scenario}}} & \multirow{2}{*}{{Metric}} & \multicolumn{4}{c}{{Method}} \\
    \cmidrule(l){3-6}
    & & {ContactFlowFilter (Ours)} & {CPF} & {MCP-EP} & {Dual-Particle only} \\
    \midrule
    \multirow{6}{*}{\parbox{2.0cm}{\centering Single-Contact\\($n_c=1$)}}
     & M-RMSE ($\mathrm{cm}$) $\downarrow$ &\bfseries 0.04 \pm 0.08      &1.84 \pm 5.27    &0.57 \pm 1.64       &0.18 \pm 0.13  \\
     & W-RMSE ($\mathrm{cm}$) $\downarrow$ &\bfseries 0.09 \pm 0.54      &1.94 \pm 5.55    &0.71 \pm 1.85       & {N/A} \\
    \cmidrule(l){2-6}
     & M-RMSE ($\mathrm{N}$) $\downarrow$  &\bfseries 0.12 \pm 0.07   &0.14 \pm 0.16    &0.13 \pm 0.07       & {N/A} \\
     & W-RMSE ($\mathrm{N}$) $\downarrow$  &\bfseries 0.17 \pm 0.07   &0.20 \pm 0.16    &0.17 \pm 0.08       & {N/A} \\
    \cmidrule(l){2-6}
     & PLS $\downarrow$          &\bfseries 5.67 \pm 8.52                &6.83 \pm 10.38   &7.95 \pm 13.34      &6.56 \pm 16.84 \\
     & EMD $\downarrow$          &\bfseries 0.70 \pm 1.84                &2.22 \pm 5.18    &1.41 \pm 1.95       &0.87 \pm 1.80 \\
    \midrule
    \multirow{6}{*}{\parbox{2.0cm}{\centering Dual-Contact\\($n_c=2$)}}
     & M-RMSE ($\mathrm{cm}$) $\downarrow$ &\bfseries 0.51 \pm 2.66      &2.94 \pm 5.92    &2.50 \pm 5.79       &0.59 \pm 1.56 \\
     & W-RMSE ($\mathrm{cm}$) $\downarrow$ &\bfseries 2.03 \pm 3.71      &4.69 \pm 6.69    &4.84 \pm 6.50       & {N/A} \\
    \cmidrule(l){2-6}
     & M-RMSE ($\mathrm{N}$) $\downarrow$  &\bfseries 2.73 \pm 3.21      &4.65 \pm 4.59    &4.29 \pm 4.06       & {N/A} \\
     & W-RMSE ($\mathrm{N}$) $\downarrow$  &\bfseries 6.09 \pm 5.03      &8.88 \pm 6.70    &9.44 \pm 6.81       & {N/A} \\
    \cmidrule(l){2-6}
     & PLS $\downarrow$         &\bfseries 9.77 \pm 14.57                &10.26 \pm 18.36  &11.80 \pm 19.14     &14.37 \pm 24.40 \\
     & EMD $\downarrow$         &\bfseries 3.29 \pm 4.33                 &6.37 \pm 7.17    &5.40 \pm 6.98       &3.48 \pm 4.08 \\
    \bottomrule
    \end{tabular*}
    \label{tab:exp_belief}
\vspace{-4mm}
\end{table*}

%% file: table/exp_real.tex
\begin{table*}[!ht]
\caption{\textbf{Experiment {\expreal}:} Quantitative results of the real-world evaluation}
\scriptsize
\centering
    \sisetup{
      table-align-uncertainty = true,
      separate-uncertainty = true,
      detect-weight = true,  
      mode = text,
    }
    \renewcommand{\arraystretch}{1.1} 
    
    \begin{tabular*}{\textwidth}{
      @{\extracolsep{\fill}} 
      l 
      l 
      S[table-format=2.3(3.2)] 
      S[table-format=2.3(3.2)] 
      S[table-format=2.3(3.2)] 
      S[table-format=2.3(3.2)] 
      @{}
    }
    \toprule
    \multirow{2}{*}{\parbox{2.0cm}{\centering {Scenario}}} & \multirow{2}{*}{{Metric}} & \multicolumn{4}{c}{{Method}} \\
    \cmidrule(l){3-6}
    & & {ContactFlowFilter (Ours)} & {CPF} & {MCP-EP} & {Dual-Particle only} \\
    \midrule
    \multirow{6}{*}{\parbox{2.0cm}{\centering Single-Contact\\($n_c=1$)}}
     & M-RMSE ($\mathrm{cm}$) $\downarrow$ &\bfseries 0.36 \pm 0.52    &1.36 \pm 3.99    &0.77 \pm 1.40    &0.51 \pm 0.63  \\
     & W-RMSE ($\mathrm{cm}$) $\downarrow$ &\bfseries 1.62 \pm 1.08    &2.88 \pm 5.48    &2.99 \pm 2.84    & {N/A} \\
    \cmidrule(l){2-6}
     & M-RMSE ($\mathrm{N}$) $\downarrow$  &\bfseries 1.98 \pm 1.11    &2.36 \pm 2.66    &2.02 \pm 1.70       & {N/A} \\
     & W-RMSE ($\mathrm{N}$) $\downarrow$  &\bfseries 2.62 \pm 2.43    &3.86 \pm 4.29    &6.17 \pm 5.91       & {N/A} \\
    \cmidrule(l){2-6}
     & PLS $\downarrow$          &\bfseries 8.85 \pm 14.03  &12.36 \pm 34.12  &30.10 \pm 34.59  &16.76 \pm 30.76 \\
     & EMD $\downarrow$          &\bfseries 1.14 \pm 1.94    &2.62 \pm 6.91    &3.14 \pm 2.39    &2.15 \pm 3.28 \\
    \midrule
    \multirow{6}{*}{\parbox{2.0cm}{\centering Dual-Contact\\($n_c=2$)}}
     & M-RMSE ($\mathrm{cm}$) $\downarrow$ &\bfseries 0.88 \pm 2.31   &5.49 \pm 11.36    &2.44 \pm 5.15    &1.04 \pm 1.59 \\
     & W-RMSE ($\mathrm{cm}$) $\downarrow$ &\bfseries 2.90 \pm 3.51   &7.13 \pm 11.24    &4.51 \pm 5.63    & {N/A} \\
    \cmidrule(l){2-6}
     & M-RMSE ($\mathrm{N}$) $\downarrow$  &\bfseries 3.25 \pm 2.78    &6.60 \pm 7.13    &4.34 \pm 4.22       & {N/A} \\
     & W-RMSE ($\mathrm{N}$) $\downarrow$  &\bfseries 6.82 \pm 4.57   &10.66 \pm 7.99    &8.98 \pm 6.66       & {N/A} \\
    \cmidrule(l){2-6}
     & PLS $\downarrow$         &\bfseries 12.52 \pm 25.41  &34.89 \pm 94.74  &19.96 \pm 56.66  &17.11 \pm 30.29 \\
     & EMD $\downarrow$         &\bfseries 3.18 \pm 4.67   &7.93 \pm 12.25    &4.75 \pm 6.29    &3.42 \pm 3.63 \\
    \bottomrule
    \end{tabular*}
    \label{tab:exp_real}
\end{table*}